\documentclass{article}
\pdfoutput=1

\usepackage{PRIMEarxiv}
\usepackage[numbers]{natbib}
\usepackage{tabularx,ragged2e,booktabs}
\usepackage[utf8]{inputenc} 
\usepackage[T1]{fontenc}    
\usepackage{hyperref}       
\usepackage{url}            
\usepackage{booktabs}       
\usepackage{amsfonts}       
\usepackage{nicefrac}       
\usepackage{microtype}      
\usepackage{lipsum}
\usepackage{float}
\usepackage{fancyhdr}       
\usepackage{graphicx}       
\graphicspath{{media/}}     

\title{Evaluating Large Language Models on the GMAT: Implications for the Future of Business Education 
}

\author{
  Vahid Ashrafimoghari , Necdet Gürkan , Jordan W. Suchow \\
  Stevens Institute of Technology\\
  \texttt{vashraf1@stevens.edu} \\
}

\begin{document}
\maketitle

\begin{abstract}
The rapid evolution of artificial intelligence (AI), especially in the domain of Large Language Models (LLMs) and generative AI, has opened new avenues for application across various fields, yet its role in business education remains underexplored. This study introduces the first benchmark to assess the performance of seven major LLMs—OpenAI's models (GPT-3.5 Turbo, GPT-4, and GPT-4 Turbo), Google's models (PaLM 2, Gemini 1.0 Pro), and Anthropic's models (Claude 2 and Claude 2.1)— on the GMAT, which is a key exam in the admission process for graduate business programs. Our analysis shows that most LLMs outperform human candidates, with GPT-4 Turbo not only outperforming the other models but also surpassing the average scores of graduate students at top business schools. Through a case study, this research examines GPT-4 Turbo's ability to explain answers, evaluate responses, identify errors, tailor instructions, and generate alternative scenarios. The latest LLM versions— GPT-4 Turbo, Claude 2.1, and Gemini 1.0 Pro— show marked improvements in reasoning tasks compared to their predecessors, underscoring their potential for complex problem-solving. While AI's promise in education, assessment, and tutoring is clear, challenges remain. Our study not only sheds light on LLMs' academic potential but also emphasizes the need for careful development and application of AI in education. As AI technology advances, it is imperative to establish frameworks and protocols for AI interaction, verify the accuracy of AI-generated content, ensure worldwide access for diverse learners, and create an educational environment where AI supports human expertise. This research sets the stage for further exploration into the responsible use of AI to enrich educational experiences and improve exam preparation and assessment methods.
\end{abstract}


\section{Introduction}
Artificial intelligence (AI) has witnessed substantial advancements over the past few years, which has facilitated its application across a spectrum of fields. These applications range from augmenting personal assistant technologies \cite{felix2018smart}, to innovating healthcare practices \cite{xu2021chatbot}, and to enriching educational methodologies \cite{zawacki2019systematic}. In healthcare, for instance, AI assists professionals by organizing patient records, analyzing diagnostic images, and identifying health issues \cite{aung2021promise}. AI applications have also been utilized in education to enhance administrative services and academic support \cite{zawacki2019systematic}. These systems have been developed to simulate one-to-one personal tutoring, helping educators in creating more effective educational settings. However, developing these systems is challenging; it involves not only content creation and design but also the refinement of feedback phrasing and dialogue strategies \cite{afzal2019development}.

The emergence of Large Language Models (LLMs), also known as Large Generative AI Models (LGAIMs), has been transformative for Natural Language Processing (NLP) tasks, demonstrating considerable promise in the spheres of education and assessment. As integral elements of AI, LLMs are adept at comprehending, producing, and interpreting human language. With the ongoing advancement of AI, it becomes imperative to scrutinize the proficiency and constraints of these models within educational frameworks. Our study investigates the efficacy of LLMs on the Graduate Management Admission Test (GMAT), an established benchmark for entry into graduate management programs. The objective of this evaluation is to elucidate the capabilities and limitations of LLMs in educational environments.

The GMAT plays a crucial role in the application process for business schools worldwide. It is designed to assess candidates' abilities in verbal and quantitative reasoning, analytical writing, and integrated reasoning, offering a thorough evaluation of their preparedness for the challenging academic environment of business school. Traditionally, preparation for the GMAT exam has involved human tutors, who deliver their services either in classroom settings or via online tutoring platforms. These tutors provide tailored guidance, practice tests, and review sessions, all aimed at helping candidates excel in the various segments of the exam.In today's educational environment, a wide array of companies and websites, such as Kaplan, Manhattan GMAT, Economist, and Magoosh, offer extensive GMAT exam preparation services. These services encompass a range of options, including self-paced online courses, live online classes, private tutoring, and comprehensive study materials. Their goal is to accommodate various learning styles and schedules, thereby making GMAT preparation more accessible and adaptable for prospective business school students.

However, the recent advancements in LLMs, such as GPT-4 Turbo, present a unique opportunity to revolutionize the GMAT preparation process. These sophisticated models have the potential to automate certain aspects of GMAT preparation, offering personalized, adaptive learning experiences that can match or even surpass traditional methods. For instance, they could provide instant feedback on practice questions, adapt the difficulty level based on the learner's progress, or offer targeted practice on weak areas. Moreover, they could be available 24/7, offering flexibility that traditional tutoring cannot match. The exploration of this potential marks an exciting frontier in the intersection of artificial intelligence and education, promising to make GMAT preparation more efficient, effective, and tailored to individual needs.

Building on the possibilities introduced by LLMs, this study seeks to answer the following research questions, each aimed at exploring the capabilities and potential applications of these models in the context of GMAT preparation:

\textbf{RQ1:} How do LLMs compare to human candidates in terms of performance when responding to the verbal and quantitative reasoning sections of the GMAT? 

\textbf{RQ2:} What potential benefits and drawbacks are associated with the use of LLMs in learning and education, especially for tutoring, exam preparation, and assessment? 

To address the research questions, we adopted a comprehensive approach, initiating with an analysis of model performance on GMAT exam questions. Our evaluation encompassed seven state-of-the-art general-purpose LLMs: GPT-3.5 Turbo, GPT-4, GPT-4 Turbo, Claude 2, Claude 2.1, PaLM 2, and Gemini 1.0 Pro, focusing on their abilities in the quantitative and verbal reasoning sections of the GMAT. Utilizing both the free and premium practice exams offered by the Graduate Management Admission Council (GMAC), we aimed to discern any effects of memorization or data leakage. The results revealed a negligible variance in performance between the free and premium exams.Notably, GPT-4 Turbo, employing zero-shot standard prompting, significantly outperformed the others, achieving an average accuracy of 85.07\% across three sets of official GMAT practice exams, compared to 74.13\% for GPT-4, 56.72\% for GPT-3.5 Turbo, 72.14\% for Claude 2.1, 60.2\% for Claude 2,  70.65\% for Gemini 1.0 Pro  and 50.75\% for PaLM 2. 

Our research goes beyond basic performance metrics to explore AI behavior in educational settings. We provide a comparative analysis of human and AI performance on the GMAT, discussing AI errors for targeted improvements. A case study highlights the qualitative behavior of GPT-4 Turbo, demonstrating its ability to articulate critical reasoning and interactively assist students by evaluating their responses and errors, and creating counterfactual scenarios. We conclude by reflecting on the potential impact of our findings on business education and professional development, addressing accuracy, fairness, and the broader implications for management practices. While recognizing the limitations of our evaluation methods, we discuss the necessary precautions and advancements for the real-world application of LLMs like GPT-4 Turbo. Despite the need for careful and comprehensive evaluation, we anticipate that AI models such as GPT-4 Turbo will have a significant and positive impact on the business sector, becoming essential tools in educational and professional settings.

\section{Related Work}

In recent years, several large pre-trained models have been developed, including GPT \cite{radford2018improving}, Bard \cite{Bard}, Claude 2 \cite{Claude2}, BERT \cite{devlin2018bert}, RoBERTa \cite{liu2019roberta}, and the widely used GPT-3 \cite{floridi2020gpt}, GPT-3.5 \cite{scao2022bloom}, and GPT-4 \cite{openai2023gpt4}. These models are based on transformer architecture \cite{vaswani2017attention} and have been pre-trained on massive datasets of text to generate human-like text, answer questions, assist in translation and summarization, and perform many NLP tasks with a single pre-training and fine-tuning pipeline. These developments mark significant milestones in the field of NLP and offer enormous opportunities for applications in research and industrial contexts. We anticipate that future advancements in AI will offer significant benefits, thus highlighting the need to explore their potential applications in education. 

LLMs have a range of applications within educational settings, enhancing student learning. Researchers have utilized these models to develop interactive learning tools, such as quizzes and flashcards, which in turn improve student engagement and facilitate knowledge acquisition \cite{dijkstra2022reading, gabajiwala2022quiz}. One of the advantages of using LLMs in education is their ability to help students complete their practice questions more efficiently \cite{kasneci2023chatgpt}.  Specifically, GPT-3 has been employed to generate multiple-choice questions and answers, enhancing reading comprehension \cite{dijkstra2022reading}. The study indicates that automated quiz generation not only reduces the workload for educators in creating quizzes manually but also acts as an effective tool for students to evaluate and deepen their knowledge throughout their study period and exam preparation. The researchers developed a method for automatically creating prompts that encourage learners to ask more substantive questions \cite{scialom2019ask}. The findings suggest that LLMs can significantly facilitate the promotion of curiosity-driven learning experiences and serve as a powerful mechanism to enhance the expression of curiosity among learners \cite{abdelghani2023gpt}.

Personalized learning is a promising application area for LLMs, considering the individual differences in learning styles and vast amount of educational data available \cite{ashrafimoghari2022big}. Researchers demonstrated the application of an advanced GPT-3 model, specifically calibrated for chemistry, to assess answers provided by students \cite{moore2022assessing}. This technology could significantly aid educators in appraising the quality and instructional value of student responses. Additionally, \cite{zhu2020effect} proposed a system that delivers AI-driven, personalized feedback within a high school science task. They found that this system effectively supported students in refining their scientific reasoning. \cite{sailer2023adaptive} observed that such feedback systems could enhance the ability of teacher trainees to substantiate educational assessments. Furthermore, \cite{bernius2022machine} noted that in large-scale courses, these systems can provide feedback on written work, potentially reducing the grading workload by 85\% while maintaining high accuracy. This enhancement also improves the perceived quality of feedback among students. Recognizing the benefits of LLMs, Khan Academy has developed an AI chatbot named Khanmigo, which functions as a virtual tutor and classroom assistant. This initiative represents a step forward in incorporating LLMs into educational platforms, aiming to enhance tutoring and coaching experiences. The goal is to provide personalized one-on-one interactions with students, demonstrating a practical application of LLMs in education \cite{khan}. Another notable example is the potential use of LLMs in foreign language learning. Their features, such as an effortless, seamless, and friendly interface, contribute to an excellent user experience. A study demonstrated that language learners appreciated the ChatGPT prompts while performing language tasks \cite{shaikh2023assessing}. In another study, researchers proposed a system that leverages LLMs to identify content that aligns with the user's interests and closely matches their proficiency level in a foreign language \cite{vlachos2023large}. 

In the domain of computer science education, GPT-3 has been utilized as an educational aid to clarify various aspects of coding, although numerous questions related to research and instructions remain open for further investigation \cite{macneil2022generating}. In a separate endeavor, \cite{bhat2022towards} have devised a method for generating evaluative questions tailored to a data science curriculum by refining a GPT-3 model with text-based educational material. These questions were evaluated for their instructions merit using both an automated categorization system employing a specifically trained GPT-3 model and expert reviews from domain specialists. The outcomes indicated that the GPT-3-generated questions were positively appraised by these experts, thus supporting the adoption of large language models in the field of data science education. In a recent study, researchers combined the capabilities of GPT-4 and GPT-3.5 to provide automated, tutor-like, high-quality programming suggestions \cite{phung2023automating}. 

In digital ecosystems for education, particularly in Augmented Reality (AR) and Virtual Reality (VR) environments \cite{ahuja2023digital, gao2021digital}, LLMs can play a transformative role. They can amplify key factors crucial for immersive user interaction with digital content. For instance, LLMs can greatly enhance the natural language processing and understanding capabilities of AR/VR systems. This enhancement enables effective and natural communication between users and the system, such as with a virtual teacher or virtual peers. The importance of this capability for immersive educational technologies was identified early as a critical aspect of usability \cite{roussou2001immersive} and is widely recognized as a key factor in improving human-AI interactions \cite{guzman2020artificial}. 

In the realm of standardized testing, \cite{ kung2023performance} examined ChatGPT's performance on the United States Medical Licensing Examination, revealing that ChatGPT's scores approached the passing threshold without the benefit of domain-specific training. These findings prompted the researchers to propose that large language models have the potential to substantially aid medical education and even contribute to clinical decision-making processes \cite{ kung2023performance}. Additionally, an investigation into ChatGPT's performance on four actual law school exams at the University of Minnesota found that it averaged a C+ grade, managing to pass in all four subjects \cite{choi2023chatgpt}. These outcomes suggest that LLMs hold promising implications for both medical and legal educational fields \cite{haupt2023ai,katz2023gpt,thirunavukarasu2023large}.

\section{Methodology}
In this study, we evaluated seven general-purpose LLMs on three GMAT exams using the zero-shot standard prompting method, as described in the following.

\subsection{Models}

\paragraph{OpenAI's GPT Family}
We tested three LLMs form OpenAI's GPT family: GPT-3.5 Turbo, GPT-4, and GPT-4 Turbo. GPT-3.5-turbo, released on March 1st, 2023, is an improved version of GPT-3.5 and GPT-3 Davinci. GPT-3.5-turbo utilizes a transformer architecture with a large number of parameters. It is trained on a diverse corpus of text data, which enhances its ability to understand and generate human-like text. This increase in scale empowers the model to tackle more complex tasks, comprehend nuanced contexts, and ultimately achieve enhanced performance \cite{ye2023comprehensive}. Training data sourced from diverse materials including books, articles, web pages, and more provides GPT-3.5-turbo broad exposure to language and contexts available across the internet. This extensive and varied training corpus allows the model to develop a multifaceted understanding of language, enhancing its capacity for remarkably human-like text generation. The last update to GPT-3.5-turbo was in September 2021, and it currently powers the freely available version of ChatGPT.

Initially launched on March 14, 2023, GPT-4 has been made accessible to the public through the paid chatbot service, ChatGPT Plus, and also via OpenAI's API. As a transformer-based model, GPT-4 employs a two-step training process. The initial pre-training phase utilizes both public data and data sourced from third-party providers to predict the subsequent token. Following this, the model undergoes a fine-tuning process, leveraging reinforcement learning feedback from both humans and AI to ensure alignment with human values and policy compliance \cite{openai2023gpt}. 

GPT-4 Turbo is the latest version of GPT and appears to be the most advanced AI model currently available in the market. According to the OpenAI's website, this enhanced version boasts increased capabilities, possesses knowledge up to April 2023, and features an expanded context window of 128k, allowing it to process the equivalent of 300 pages of text in one prompt. Furthermore, it offers a cost reduction, with input tokens being 3 times more affordable and output tokens 2 times more affordable than those of the original GPT-4 model. Additionally, this model can generate up to 4096 output tokens \cite{GPT-4-Turbo}.

While both GPT-4 and GPT-4 Turbo possess multi-modal functionalities \cite{openai2023gpt4,GPT-4V}, our research focuses exclusively on the text-based versions of these models. Henceforth, any reference to GPT-4 or GPT-4 Turbo within this document will specifically denote the text-only versions without visual processing features.

\paragraph{Anthropic's Claude Family}
Anthropic, the firm responsible for Claude, was established in 2021 by a team of former OpenAI employees who contributed to the development of OpenAI's GPT-2 and GPT-3 models.  In early 2023, after conducting a closed alpha with a select group of commercial partners, Claude's model was incorporated into products such as Notion AI, Quora's Poe, and DuckDuckGo's DuckAssist. In March 2023, Claude expanded its API access to a broader range of businesses. Subsequently, in July 2023, it launched its chatbot to the public, coinciding with the release of the Claude 2 model. While Claude 2 may not yet match the capabilities of GPT-4, it is rapidly advancing and consistently outperforms most other AI models in standardized tests. Claude 2 distinguishes itself with its capacity to manage up to 100K tokens per prompt, equivalent to approximately 75,000 words. This is a twelve-fold increase compared to the standard amount offered by GPT-4. Also, while GPT's knowledge cutoff is September 2021, Claude 2 benefits from training data extending up to early 2023\cite{Claude2}. Anthropic unveiled Claude 2.1 on November 23, 2023, marking the newest update to their LLM lineup. As stated on their website, this latest version brings enhancements crucial for enterprise applications, such as a top-tier context window capable of handling 200K tokens, marked improvements in reducing instances of model-generated inaccuracies, and system prompts. Additionally, Claude 2.1 introduces a beta feature known as 'tool use.' Alongside these advancements, Anthropic has revised their pricing structure to increase cost-effectiveness across all their model offerings \cite{claude-2.1}. Both Clude 2.0, and Claude 2.1 are accessible through the claude.ai, Anthropic's LLM-based chatbot.

\paragraph{Google's LLM Families}

The Pathways Language Model (PaLM) is a LLM with 540 billion parameters, constructed using the Transformer architecture and developed by Google AI. The subsequent iteration, PaLM 2, was released in late 2022 and employs dynamic pathway selection during inference to determine the most suitable pathway for a given input \cite{anil2023palm}.  Bard, Google's conversational AI agent introduced in March 2023, was initially based on the LaMDA family of LLMs and later integrated PaLM 2\cite{Bard}.  
The latest generation of Google's LLMs is the Gemini Family. Developed from the Transformer architecture, Gemini models incorporate architectural improvements and optimization techniques for stable, scalable training and efficient inference on Google's Tensor Processing Units. These models are capable of handling a context length of 32k, employing efficient attention mechanisms such as multi-query attention(\cite{team2023gemini}). The first release, Gemini 1.0, is available in three main sizes—Ultra, Pro, and Nano—to accommodate a variety of applications. Currently, Gemini 1.0 Pro, a cost and latency-optimized version of Gemini 1.0, has been integrated into the new Bard. Google has announced plans to make Gemini 1.0 Ultra accessible through Bard Advanced in early 2024  \cite{new-bard}. For this study, we analyzed data from the legacy Bard tested between July 16 and October 2, 2023, and results obtained from Bard after December 6, 2023, for Gemini 1.0 Pro. While Gemini models offer multi-modal capabilities \cite{team2023gemini}, our research is concentrated on the text-only variant of Gemini 1.0 as implemented in the new Bard. Moving forward, any mention of Gemini 1.0 Pro in this paper will refer specifically to the text-only iteration, excluding visual processing features.

\subsection{Graduate Management Admission Test (GMAT)}

The GMAT exam serves as a differentiating factor for more than 100,000 candidates every year during the business school admissions process, with a significant 90\% of new MBA enrollments relying on a GMAT score \cite{GMAT}. The GMAT exam is recognized as a reliable indicator of a student's potential for success. Each GMAT exam consists of 31 quantitative reasoning, 36 verbal reasoning, 12 integrated reasoning problems, and an analytical writing essay. The GMAT Total Score is derived from the Verbal and Quantitative sections of the exam. This Total Score is based on the test taker's calculated performance before the scores for the Quantitative Reasoning and Verbal Reasoning sections are assigned. This raw calculation is then converted into a number within the Total Score range of 200 to 800. As only the performance in the quantitative and verbal reasoning sections contributes to the total score, we focused solely on the performance in these two sections, disregarding the performance in Integrated Reasoning and Analytical Writing. The quantitative reasoning section comprises two tasks: Problem Solving and Data Sufficiency. The verbal reasoning section is also divided into three tasks: Reading Comprehension, Critical Reasoning, and Sentence Correction. 

In this study, we deployed seven different LLMs—GPT 3.5-turbo, GPT-4, GPT-4 Turbo, Claude2, Claude 2.1, PaLM 2, and Gemini 1.0 Pro—to generate responses for three official GMAT practice exams. We emphasize that the materials used in this study were officially acquired and sourced from the Graduate Management Admission Council (GMAC), the authoritative body that oversees and administers the GMAT examination process. Given that the GMAT exam employs an adaptive testing format—wherein a test taker's performance on prior questions dictates the difficulty level of subsequent ones—we maintain that administering the tests through the official GMAT exam website constitutes the most authentic method for evaluating the performance of LLMs. This approach is likely to provide the most accurate estimation of the AI's proficiency and competency. By leveraging the official testing platform, we can ensure that the AI is subjected to the same dynamic adjustments in difficulty that human test-takers face, thereby offering a fair and rigorous assessment of its capabilities in a standardized context. Our choice of exams was strategic; to mitigate the potential influence of publicly available free practice exams on the results, we included premium practice exams among the three exam sets for each model. These premium exams were not publicly accessible, thereby ensuring a level of novelty and challenge for the LLMs.

\subsection{Prompting}

\begin{figure}[ht]
    \centering
    \includegraphics[width=1\linewidth]{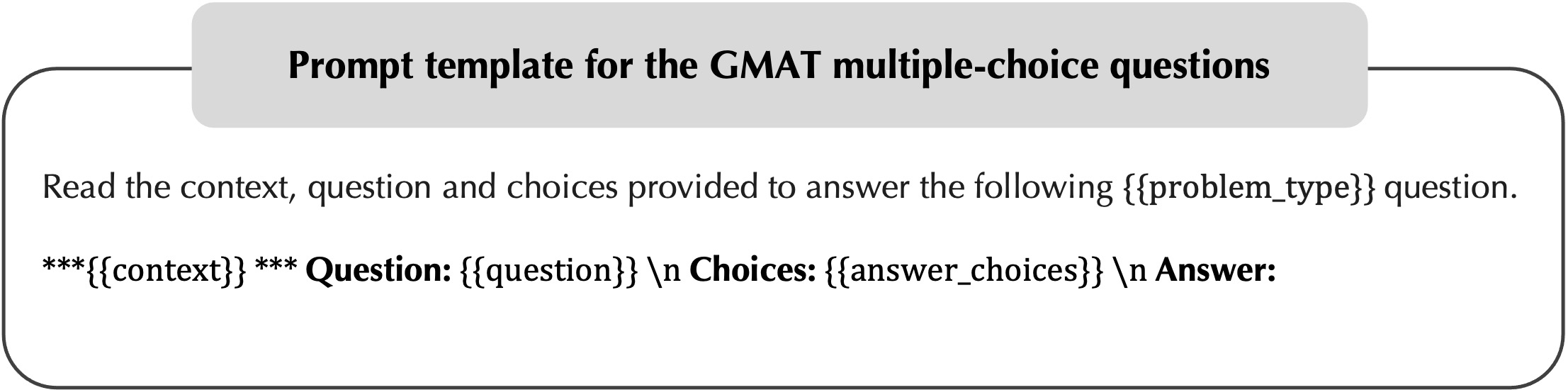}
    \caption{The template employed for generating prompts for every multiple-choice question. Elements shown in double braces {{}} are substituted with question-specific values.}
    \label{fig:template}
\end{figure}

\begin{figure}[H]
    \centering
    \includegraphics[width=1\linewidth]{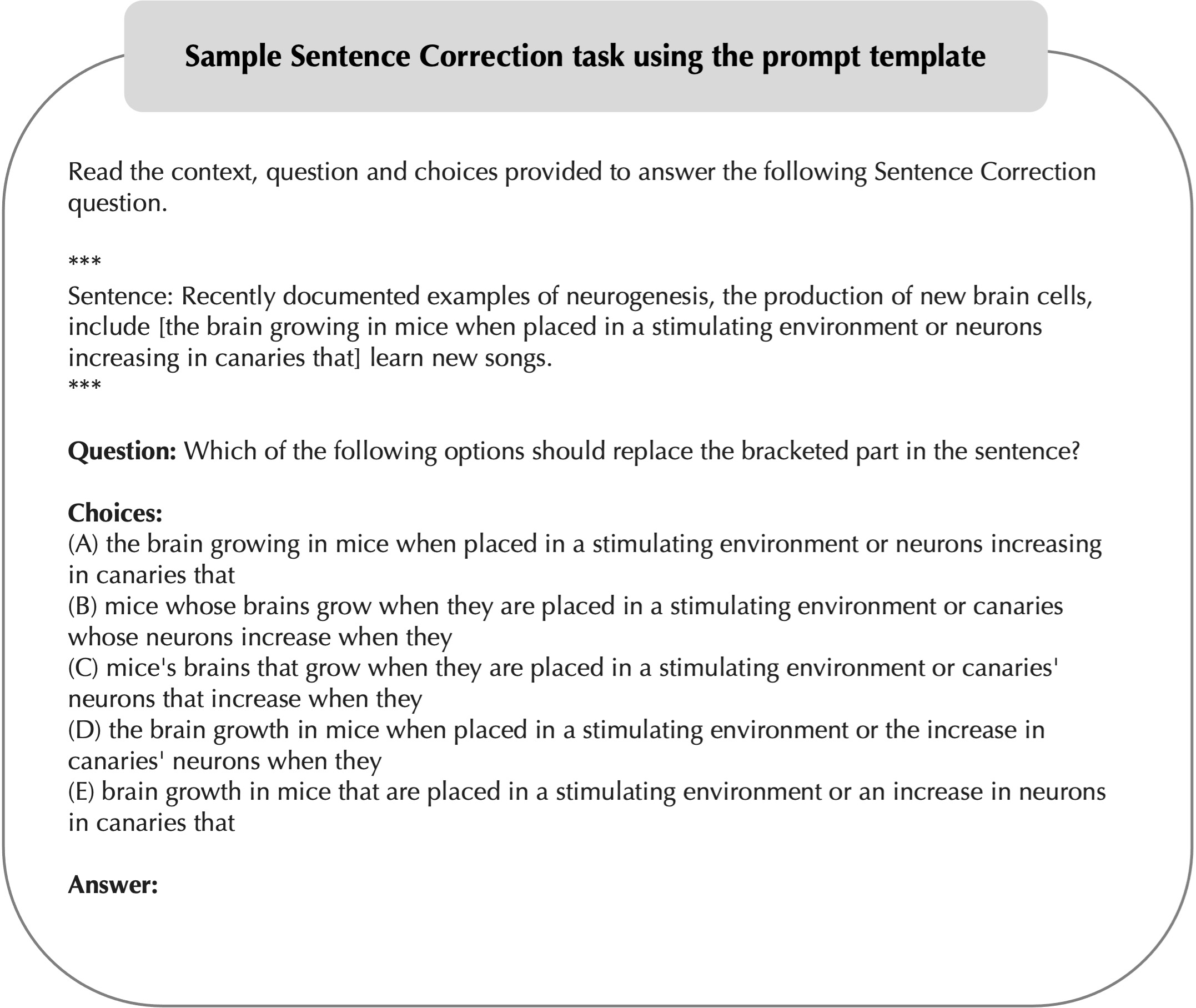}
    \caption{An example of implementation of template shown in from Figure \ref{fig:template}.}
    \label{fig:sample}
\end{figure}

\begin{figure}[H]
    \centering
    \includegraphics[width=1\linewidth]{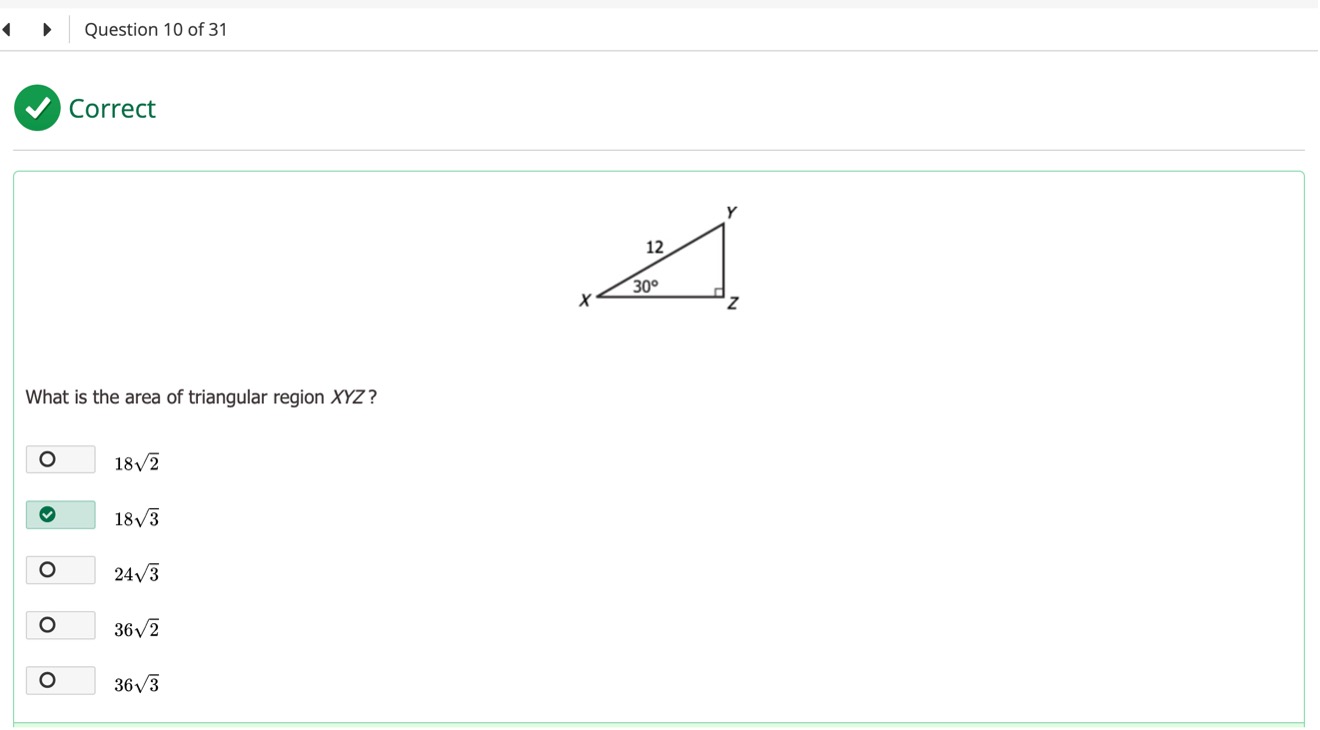}
    \caption{Original problem statement with image}
    \label{fig:original}
\end{figure}

\begin{figure}[H]
    \centering
    \includegraphics[width=1\linewidth]{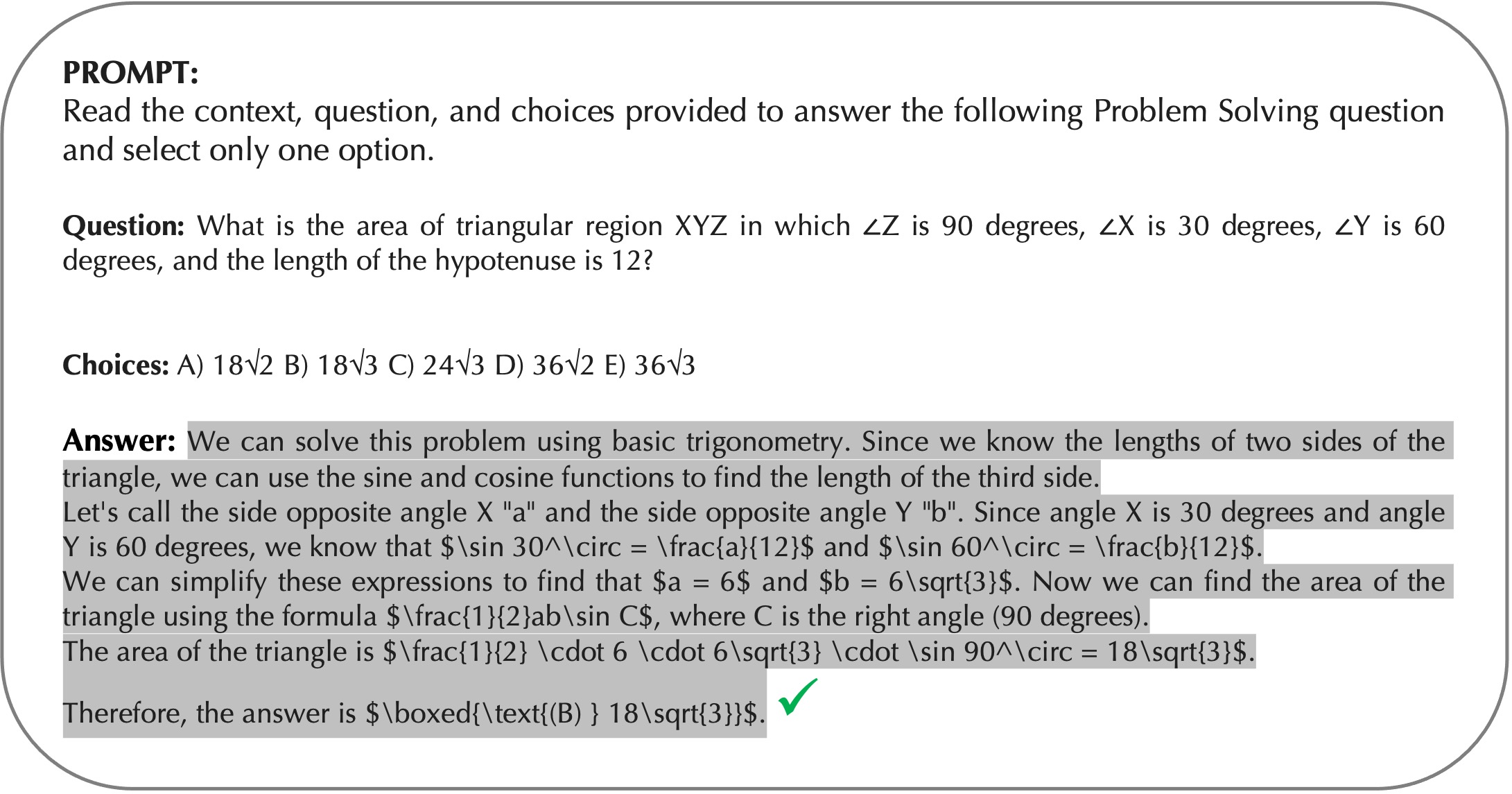}
    \caption{The prompt describes the problem shown in Figure \ref{fig:original}, presented without an accompanying image. The correct answer provided by GPT-3.5 Turbo is highlighted in grey.}
    \label{fig:convert}
\end{figure}

To establish a baseline for model performance and to ensure fair comparison with human participants, we applied a simple zero-shot prompting technique for the models being evaluated. Additionally, to reduce the potential for in-context learning during chat-based interactions, we posed each question in a separate chat window using LLM-based chatbots, rather than within a continuous conversation thread. This precaution ensured that each model's response was independent and not influenced by previous interactions, providing a more accurate measure of standalone performance. We standardized the prompt template, as illustrated in Figure \ref{fig:template}, customizing it for different problem types, with a complete example shown in Figure \ref{fig:sample}.   This approach maintained prompt consistency across similar problems, allowing for automated model responses with minimal human intervention, limited to entering the problem content and prompt.
In our study, we focused exclusively on the text-based capabilities of LLMs. Consequently, for questions originally accompanied by images, we converted them into descriptive math word problems (Figures \ref{fig:original} and \ref{fig:convert}). This translation process ensured that the essence of the visual information was accurately captured and conveyed through text, allowing the LLMs to process and respond appropriately.

\section{Results and Discussion}
\subsection{Performance of LLMs on the GMAT}

Table \ref{tab:sections_tasks} presents an overview of the average performance of LLMs on the GMAT exam, both overall and across different tasks. It is worthy to note that the overall average mentioned in the table is calculated by dividing the number of correct answers by the total number of questions across all three exam sets for each model.

\begin{table}[h!]
    \centering
    \begin{tabular}{l >{\centering\arraybackslash}p{1cm} >{\centering\arraybackslash}p{1cm} >{\centering\arraybackslash}p{1.2cm} >{\centering\arraybackslash}p{1.5cm} >{\centering\arraybackslash}p{1.25cm} >{\centering\arraybackslash}p{1.65cm} >{\centering\arraybackslash}p{2cm}}
    \hline
    \toprule
    Sections \& Tasks&  \textbf {GPT-4 Turbo} &  GPT-4 &  GPT-3.5 Turbo& Claude 2.1&  Claude 2&  Gemini Pro (New Bard)& PaLM 2 (Legacy Bard)
\\\hline
\textbf{Quantitative Reasoning} 
& \textbf{74.19} 
&  64.52 
&  53.76
& 61.29
& 46.24
& 68.82
& 44.09
\\ \hline
Data Sufficiency
&  \textbf{60.98}
&  56.10
&  39.02
& 56.10
& 36.59
& 60
& 32.50
\\
Problem Solving
&  \textbf{84.62}
&  71.15
&  64.15
& 65.38
& 53.85
& 75.47
& 52.83
\\ \hline
\textbf{Verbal Reasoning}
&  \textbf{94.44}
&  82.41
&  61.11
& 81.48
& 72.22
& 72.22
& 56.48
\\ \hline
Reading Comprehension
&  \textbf{100}
&  97.44
&  70.73
& 100
& 97.44
& 87.50
& 79.49
\\
Critical Reasoning
&  \textbf{96.30}
&  81.48
&  55.56
& 75
& 64.29
& 66.67
& 50
\\
Sentence Correction
&  \textbf{87.80}
&  69.05
&  55.00
& 68.29
& 53.66
& 60.98
& 39.02
\\
\bottomrule
\textbf{Overall Average}&  \textbf{85.07}&  74.13 &  57.71 & 72.14 & 60.20 & 70.65 & 50.75\\
\bottomrule
    \end{tabular} 
\\
    \caption{The comparison of model performance across various GMAT sections and tasks reveals that GPT-4 Turbo outperforms the other models in all sections and tasks.}
    \label{tab:sections_tasks}
\end{table}

Table \ref{tab:sections_tasks} demonstrates that across the board, all models exhibit their highest proficiency in the reading comprehension task. On the other hand, they struggle the most with the data sufficiency task, which emerges as their Achilles' heel. For the remaining tasks—problem-solving, critical reasoning, and sentence correction—the models' performance falls within the average range, indicating a balanced level of proficiency in these areas but without the peaks observed in reading comprehension or the troughs seen in data sufficiency.

\paragraph{Quantitative reasoning}
When mathematical problems have unambiguous presentations and sequential solutions, LLMs can apply their extensive computational power to reach correct answers quickly and on vast scales unmatched by humans \cite[See,][]{lewkowycz2022solving}. However, despite exceeding human performance on well-specified calculations, LLMs still lack deeper comprehension of the mathematical concepts themselves. Instead, these models depend fundamentally on pre-defined algorithms. As a result, they fluctuate when problems require creative approaches, judgments akin to human intuition, or the ability to integrate multiple abstract concepts. Vague or complex multi-step problems strain LLMs even further. 
\paragraph{Reading comprehension}
Previous research has shown pre-trained language models are typically excel at summarizing, classifying, and paraphrasing texts, demonstrating a keen ability to identify the main ideas within passages \cite[e.g.,][]{min2023recent}. Therefore, their consistently high performance in tackling reading comprehension problems is to be expected. This proficiency can be attributed to their advanced language processing capabilities, which allow them to effectively analyze and interpret textual information. Consequently, they can accurately understand the context, draw inferences, and provide precise responses to comprehension questions. However, LLMs struggle with abstract or complex concepts that require a level of reasoning or inference beyond the literal text \cite[e.g.,][]{qiu2023phenomenal,zhou2023complementary}.
\paragraph{Critical reasoning}
LLMs are adept at pattern recognition \cite{jin2023time} and inference-making \cite{guo2023leveraging}, skills that serve them well in answering critical reasoning questions focused on identifying assumptions and assessing arguments. Yet, they can falter with subtle or complex logical reasoning, especially when questions demand an understanding of new evidence or hypothetical situations, as LLMs depend on data-driven patterns rather than true understanding or practical knowledge.
\paragraph{Sentence correction}
Ultimately, LLMs, pre-trained on an extensive set of grammar rules, are proficient at identifying a broad spectrum of grammatical errors. However, they often stall when errors require an understanding of context or language nuances. Complex sentences, particularly those with multiple clauses or unconventional structures, can pose challenges \cite{veres2022large}. LLMs' reliance on programmed rules can lead to oversights in recognizing exceptions or stylistically acceptable rule deviations. Additionally, LLMs sometimes cannot grasp deeper meanings or implications of information and can struggle with context, especially when it involves cultural, historical, or personal references \cite{chang2023survey}.

\subsection{Performance of LLMs versus human candidates}

\begin{figure}[H]
    \centering
    \includegraphics[width=1\linewidth]{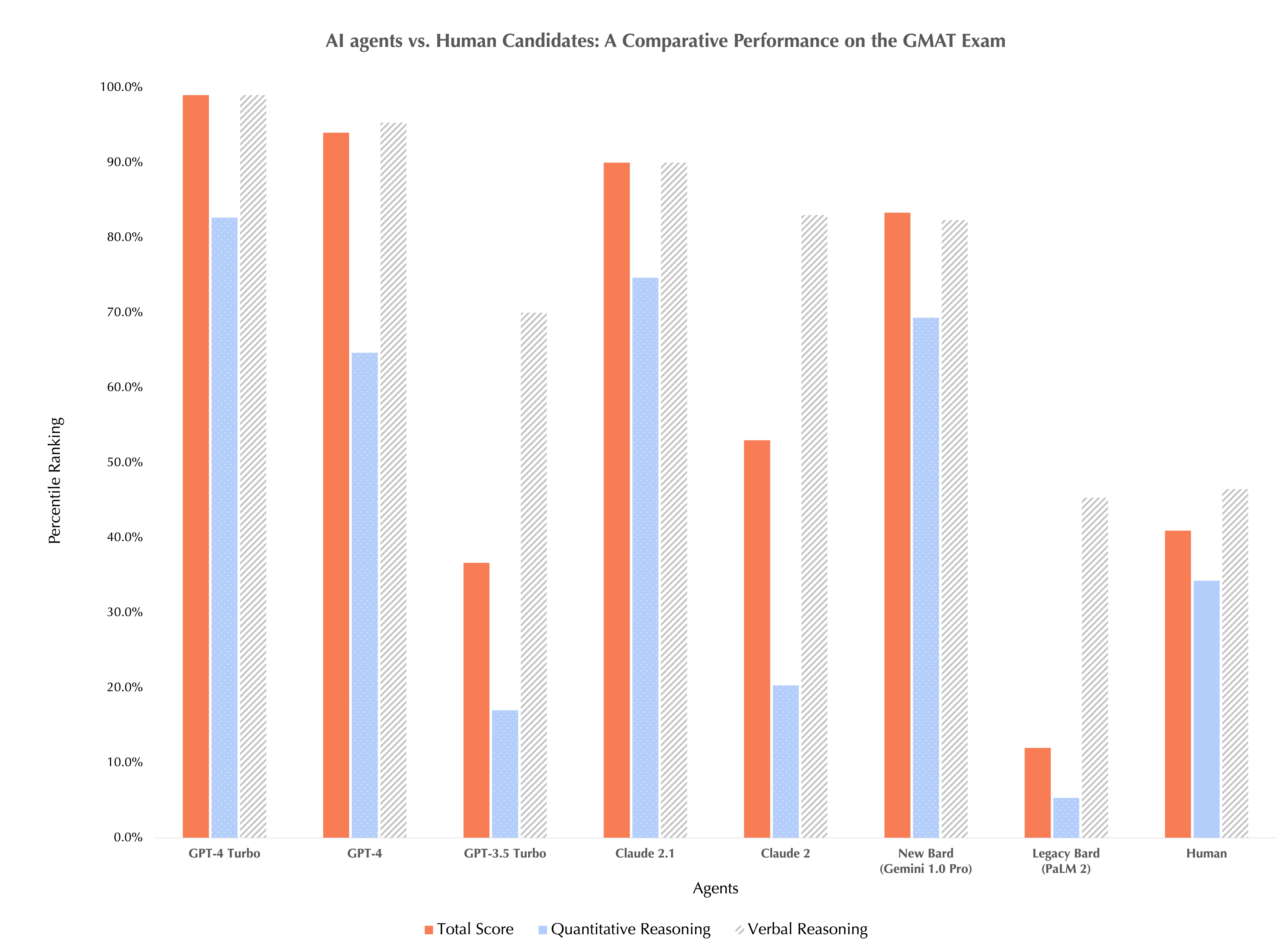}
    \caption{This figure presents a comparative analysis of the average performance across seven LLMs and human candidates in quantitative reasoning, verbal reasoning, and total GMAT scores.}
    \label{fig: agents}
\end{figure}

Figure \ref{fig: agents} illustrates the comparative performance of various AI agents, representing different LLMs, against human participants in the GMAT exam. On average, the LLMs attained a total score of 643.5 across all administered exams, a score that would place them above approximately 63\% of human candidates. Notably, GPT-4 Turbo leads the pack, outshining all other AI agents and the average human candidate, securing a spot within the top 1\% of test-takers based on the total exam score. Trailing GPT-4 Turbo, the other AI agents—GPT-4, Claude 2.1, the new Bard (Gemini 1.0 Pro), Claude 2, GPT-3.5 Turbo, and the legacy Bard (PaLM 2)—ranked successively, surpassing 94\%, 90\%, 83.3\%, 53\%, 36.7\%, and 12\% of human test-takers, respectively. Echoing the pattern observed in human candidates, all AI agents exhibited stronger results in verbal reasoning over quantitative reasoning. The extent of this performance gap between the two sections, however, showed variation among the AI agents. Interestingly, the most recent versions of LLMs, such as the new Bard (Gemini 1.0 Pro), Claude 2.1, and GPT-4 Turbo, closely emulated human test-taker performance patterns. The discrepancy between their verbal and quantitative scores was nearly identical to the average human candidate's spread. This alignment suggests that the latest LLMs are approaching human-like balance in handling the diverse skill sets required by the GMAT. It is important to note that the data regarding human test-takers is derived from a sample of 282,098 individuals who sat for the GMAT between January 2020 and December 2022, as reported in the GMAT score distribution \cite{Score}.

\begin{figure}[H]
    \centering
    \includegraphics[width=1\linewidth]{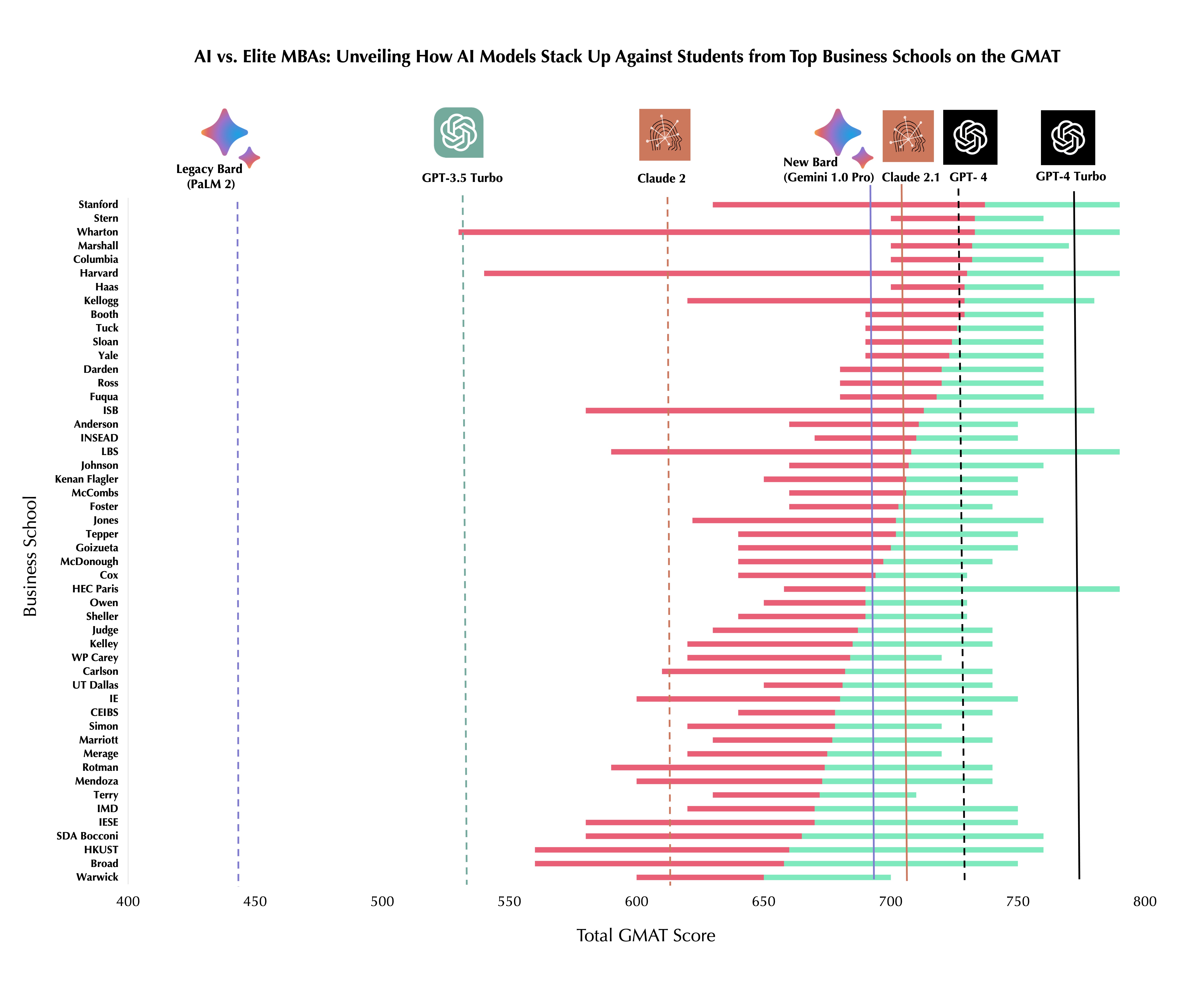}
    \caption{This graph illustrates a side-by-side comparison of average GMAT scores from seven AI models, denoted with solid lines for the latest models and dashed lines for legacy versions, against the average, minimum, and maximum GMAT scores recorded by students at the top 50 business schools, according to the Financial Times 2023 rankings\cite{FT}. The data originates from MBA class profiles publicly available on the schools' official websites. Inclusion criteria required schools to offer comprehensive information in their profiles, with the compilation organized by ascending average GMAT scores}
    \label{fig:B-schools}
\end{figure}

As depicted in Figure \ref{fig:B-schools}, AI agents' performance on the GMAT exam positions them as formidable contenders for admission into elite business schools. According to recent admissions data from the top 50 business schools, GPT-4 Turbo stands a high chance of being accepted into all these esteemed institutions. Remarkably, only seven business schools—Stanford Graduate School of Business, The Wharton School of the University of Pennsylvania, Harvard Business School, London Business School, Kellogg School of Management at Northwestern University, HEC Paris, and Indian School of Business—boast applicants with GMAT scores that exceed those of GPT-4 Turbo. This highlights GPT-4 Turbo's exceptional performance, which not only competes with but also often exceeds that of the majority of human candidates at these prestigious institutions. Moreover, GPT-4's GMAT results align with or surpass the average applicant scores at the majority of the top 50 business schools, implying a strong likelihood of GPT-4 gaining admission to nearly all these programs. The average GMAT scores of MBA classes at only the top 9 business schools exceed GPT-4's scores, underscoring its potential for acceptance. Claude 2.1 and the new Bard, despite having nearly identical GMAT scores, perform above the average of applicants to almost half of the top 50 business schools. This indicates that these AI agents are competitive applicants for many top-tier programs. Conversely, AI agents based on older LLM versions—Claude 2, GPT-3.5 Turbo, and legacy Bard—have a tiny chance of admission to the top 50 business schools, as their average GMAT scores significantly trail behind the applicant averages for these institutions.

\subsection{Limitations of LLMs}

\begin{figure}[H]
    \centering
    \includegraphics[width=1\linewidth]{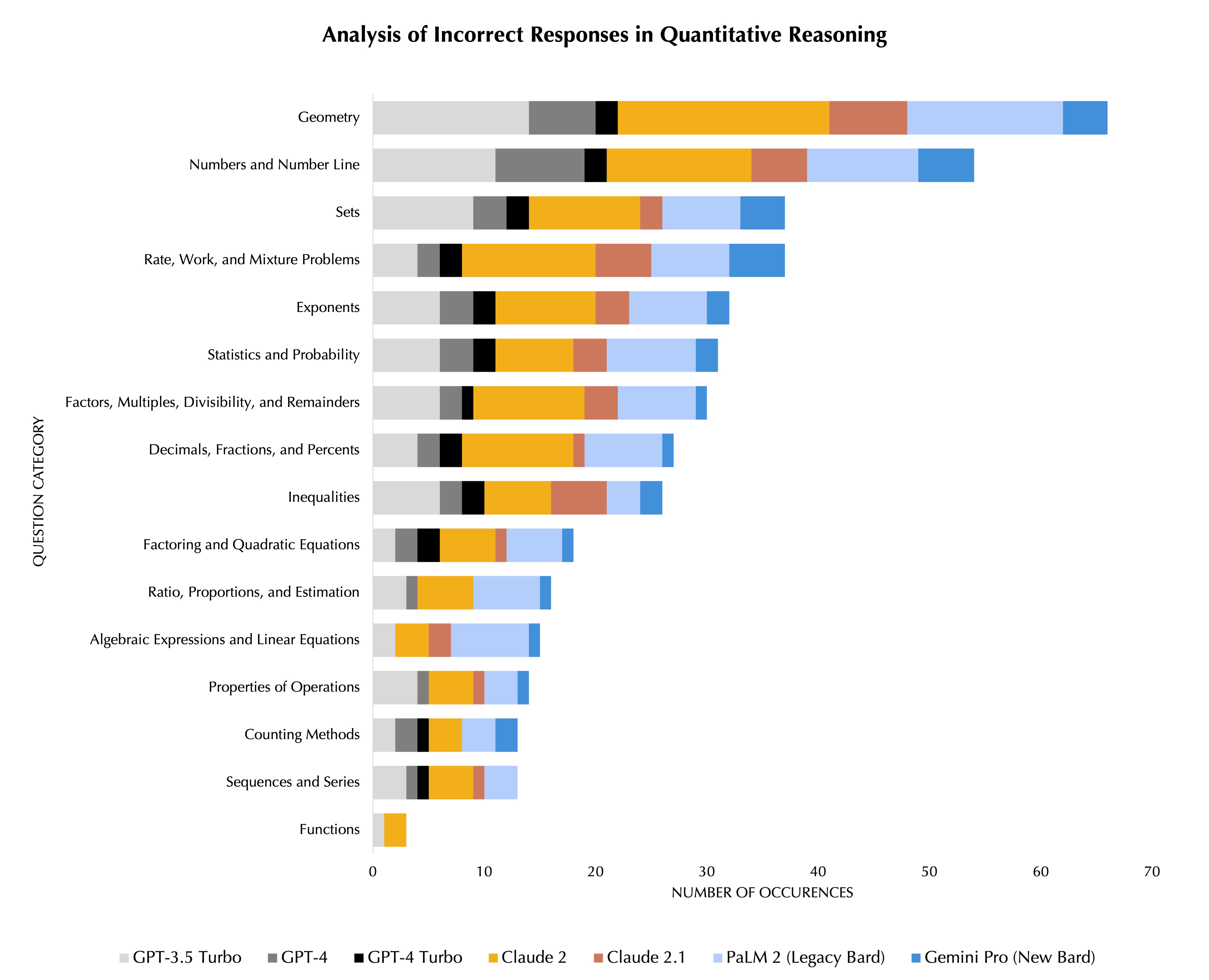}
    \caption{A detailed categorization of LLM errors in Quantitative Reasoning  for targeted improvement}
    \label{fig:Q_errors}
\end{figure}

\begin{figure}[H]
    \centering
    \includegraphics[width=0.7\linewidth]{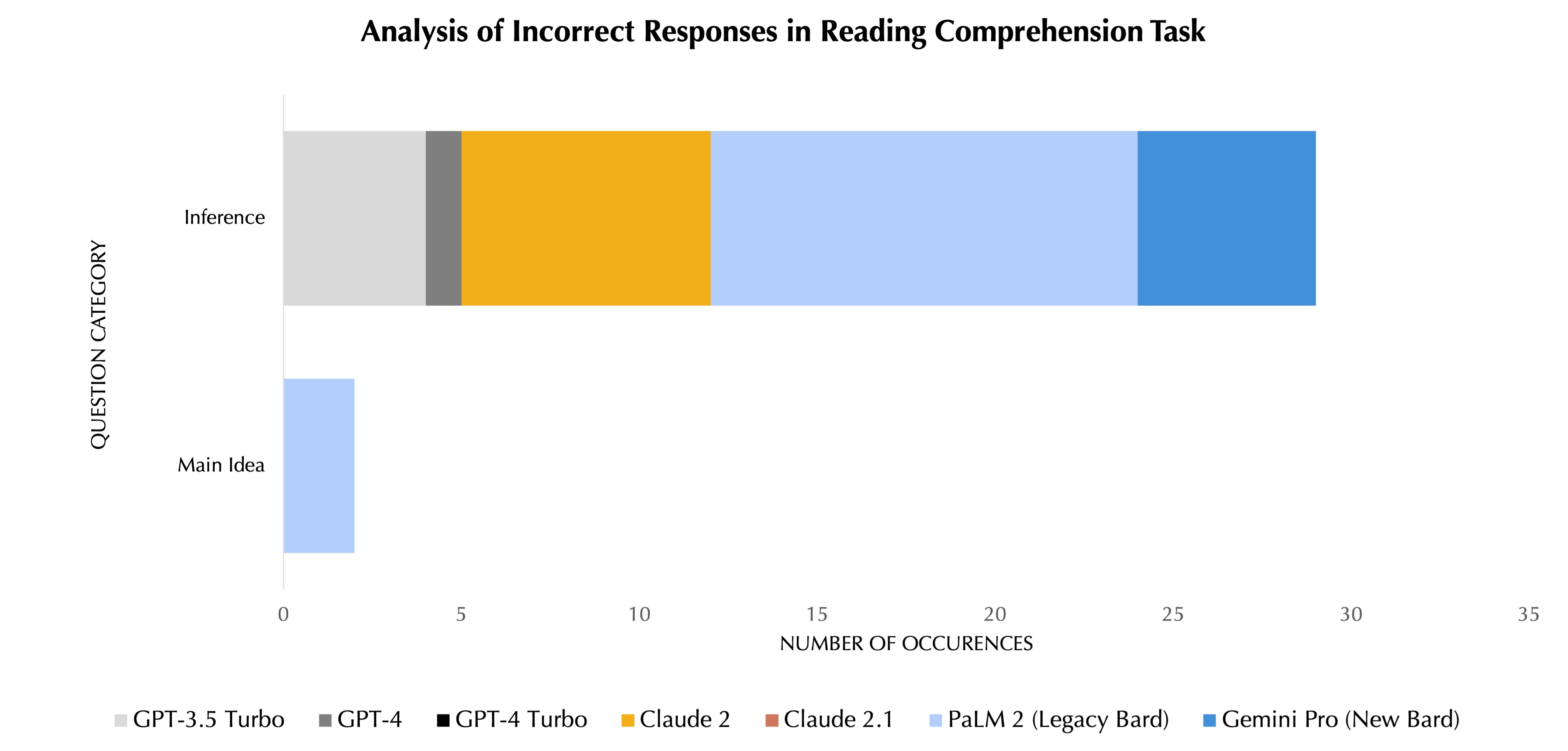}
    \caption{A detailed categorization of LLM errors in Reading Comprehension for targeted improvement}
    \label{fig:RC_errors}
\end{figure}

\begin{figure}[H]
    \centering
    \includegraphics[width=1\linewidth]{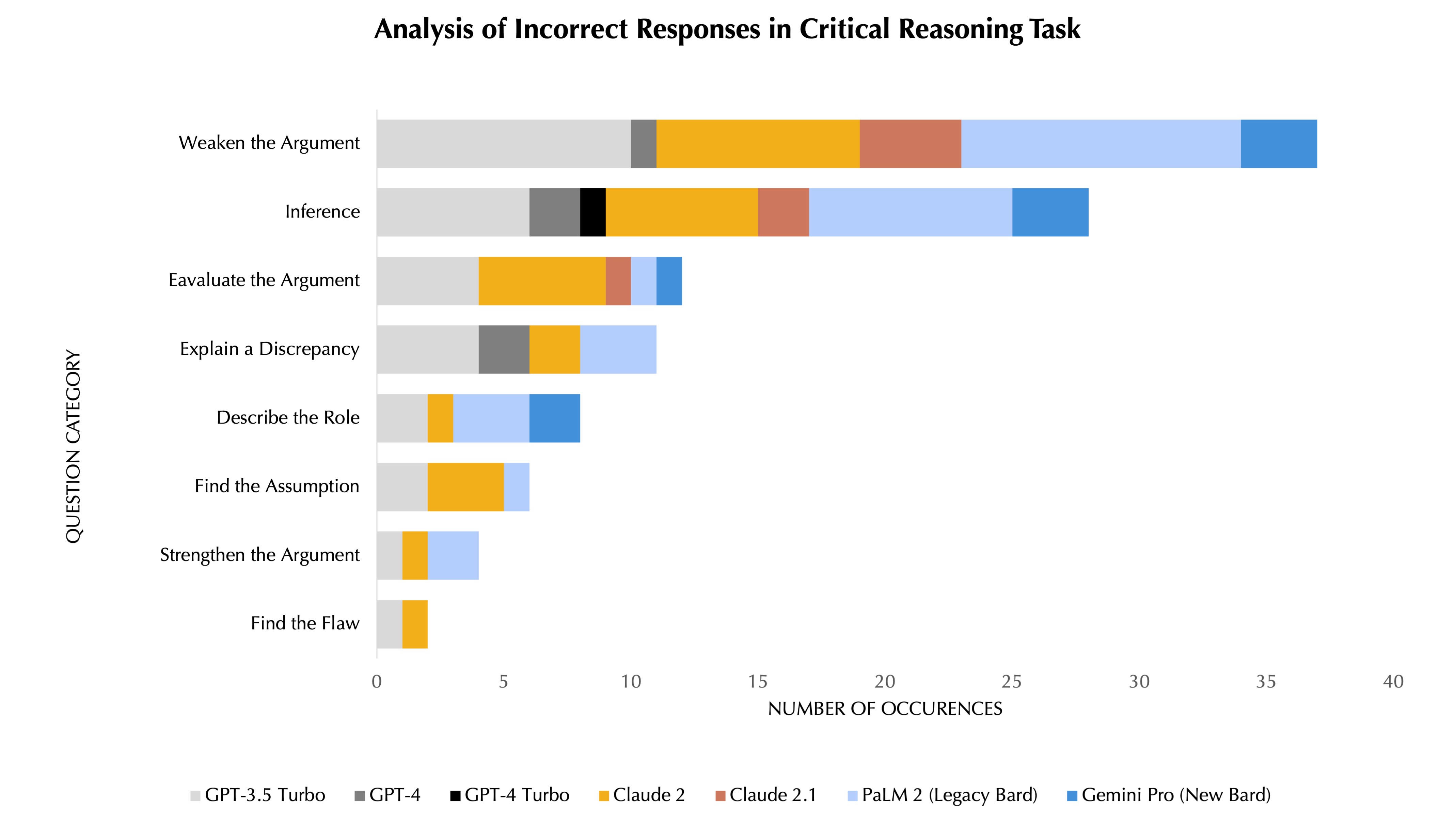}
    \caption{A detailed categorization of LLM errors in Critical Reasoning for targeted improvement}
    \label{fig:CR_errors}
\end{figure}

\begin{figure}[H]
    \centering
    \includegraphics[width=1\linewidth]{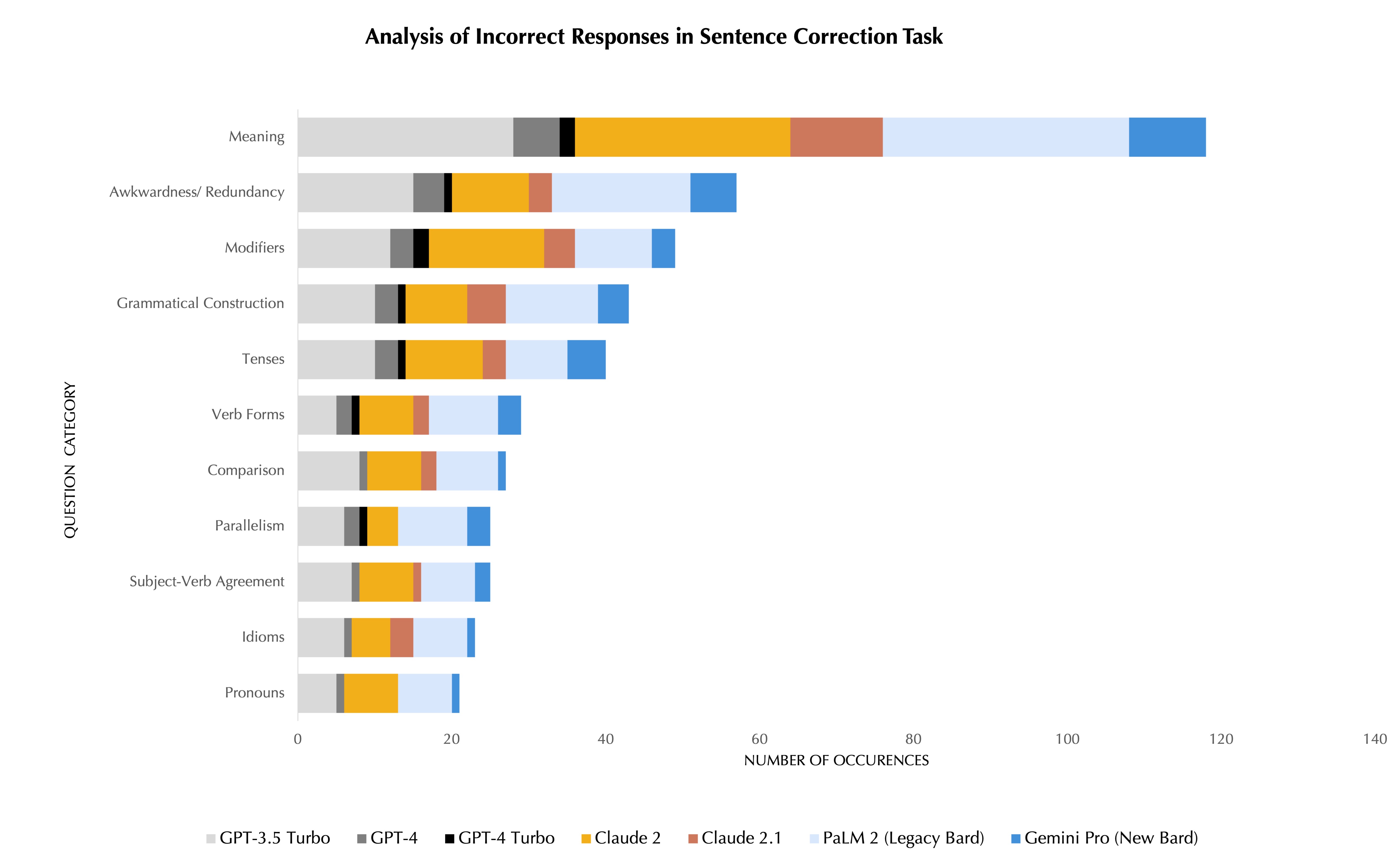}
    \caption{A detailed categorization of LLM errors in Sentence Correction for targeted improvement}
    \label{fig:SC_errors}
\end{figure}

To gain a more comprehensive understanding of the limitations of LLMs, it's crucial to conduct an in-depth analysis of their incorrect responses. Figures \ref{fig:Q_errors}, \ref{fig:RC_errors}, \ref{fig:CR_errors}, and \ref{fig:SC_errors} offer a more granular perspective on the areas where LLMs struggle. These figures specifically highlight their shortcomings in solving problems in quantitative reasoning, reading comprehension, critical reasoning, and sentence correction. By examining these areas of weakness, we can better understand the challenges faced by LLMs and work towards improving their performance in these specific domains. 

A closer examination of quantitative reasoning problems answered incorrectly by LLMs reveals 16 distinct error categories, each corresponding to specific mathematical concepts. These categories include geometry, numbers and the number line, sets, factors, multiples, divisibility and remainders, decimals, fractions and percents, exponents, statistics and probability, rate, work and mixture problems, ratio, proportions and estimations, counting methods, inequalities, factoring and quadratic equations, algebraic expressions and linear equations, properties of operations, sequences and series, and functions. This classification provides a granular view of the mathematical domains that challenge LLMs, offering insights for targeted improvement. For example, latest versions of language model families have shown a reduction or elimination of errors in certain categories. GPT-4 Turbo, in comparison to earlier models, has achieved flawless performance in areas such as Ratio, Proportions, and Estimation, and Properties of Operations. Similarly, Claude 2.1 flawlessly answers Counting Methods questions, and Gemini 1.0 Pro has a perfect record in Sequences and Series, a category where other models still falter. Overall, more than half of the errors made by LLMs are concentrated in five categories: Geometry, Numbers and Number Line, Sets, Rate, Work, and Mixture Problems, and Exponents. This pattern suggests that LLMs' mistakes likely arise from challenges of processing spatial, numerical, and logical information through text. Geometry requires visualizing diagrams, which is challenging for LLMs with text alone. Numerical concepts like absolute value and ordering can be subtle and easily misinterpreted by LLMs. Set theory's logical operations, especially in complex sets, are difficult for LLMs to process accurately. Rate and mixture problems' multi-step ratios, as well as the rules of exponentiation in exponent problems, often lead to errors when LLMs fail to apply them correctly.  

Reviewing the errors made by LLMs in reading comprehension, it is obvious that they struggle with inference questions, which require the deduction of ideas that are not explicitly stated in a passage. Additionally, they occasionally have difficulty comprehending the main idea of passages. Similarly, incorrect responses to critical reasoning problems, which present a challenge for LLMs, can be categorized into eight distinct categories. These categories include 'Weaken the Argument,' which entails identifying information that weakens the author's argument; 'Inference,' which requires identifying something that must be true based on the given information; 'Find the Assumption,' which involves identifying an assumption that is not explicitly stated; 'Explain a Discrepancy,' which requires identifying information that resolves an apparent paradox in the argument; 'Evaluate the Argument,' which involves identifying information that would help determine the argument's soundness; 'Strengthen the Argument,' which requires identifying information that supports the author's argument; 'Describe the Role,' which involves identifying the roles or labels of the underlined portions of the argument; and 'Find the Flaw,' which requires identifying an illogical element in the argument. Among these categories, questions falling under 'Weaken the Argument' are the most challenging for LLMs to answer, likely because it involves understanding the author's perspective and reasoning, as well as critically evaluating the information presented. LLMs may find it difficult to recognize the weaknesses in an argument or may struggle to differentiate between information that strengthens or weakens the argument. Therefore, these types of questions pose a particular challenge for LLMs in their critical reasoning skills.

Finally, the grammatical errors that LLMs sometimes fail to identify in sentence correction questions can be grouped into 11 categories. These include errors related to meaning, modifiers, subject-verb agreement, pronouns, grammatical constructions, awkwardness or redundancies, verb forms, tenses, comparison, parallelism, and idioms. Among these, the majority of incorrect responses are due to the LLMs' inability to detect errors in meaning. This particular challenge may stem from the subtleties and variations of language that require a deep understanding of context and intent. Errors in meaning can often involve complex logical structures or subtle shifts in tone that LLMs are not always equipped to handle. Modifiers, also, must be placed correctly to avoid ambiguity, while subject-verb agreement requires careful attention to ensure that subjects and verbs match in number. Pronouns must refer to the appropriate noun, and grammatical constructions must be consistent and logical. Awkwardness or redundancies in language can make sentences less clear and concise, which LLMs might overlook. Verb forms and tenses must be used correctly to convey the proper time frame of actions, while comparisons require a balanced and accurate assessment of similarities or differences. Parallelism is essential for maintaining a consistent structure in lists and comparisons, and idioms must be used appropriately to convey the intended meaning in a culturally specific context. In sum, the complexity of these grammatical rules and the intricacies of their application occasionally makes sentence correction a challenging area for LLMs.

\section{Study Limitations and Future Directions }

Our paper demonstrates the remarkable capability of general purpose LLMs in responding to quantitative and verbal reasoning questions on the GMAT exam. In the following, we will discuss the limitations and possible expansions of our research for the future.

\subsection{Prompting Method}
Our objective in this study is to establish a benchmark for the foundational performance of LLMs when tasked with answering GMAT multiple-choice questions. We aim to achieve this using a straightforward methodology, deliberately avoiding the use of more intricate techniques such as chain-of-thought (CoT) prompting \cite{wei2022chain}, Retrieval Augmented Generation (RAG) \cite{lewis2020retrieval} or prompt chaining strategies \cite{wu2022promptchainer}. Previous research has demonstrated that these advanced methods significantly improve the capabilities of LLMs when addressing complex queries across various fields \cite{levonian2023retrieval,lievin2022can}. Moreover, the potential exists for the development of new prompting methods specifically tailored to enhance the performance of advanced language models like GPT-4 Turbo, which have not yet been fully explored or identified. Given the complexity and adaptability of such models, it stands to reason that a deliberate and systematic examination of various prompting techniques, coupled with strategic fine-tuning, has the capacity to yield significant gains in performance outcomes. This could involve experimenting with different types of prompts, varying the complexity and structure of the language used, or even customizing prompts to align more closely with the model's training data. While the pursuit of achieving the highest possible scores on benchmarks is a valid endeavor, it is not the sole focus of this paper. Our aim extends beyond simply pushing the limits of benchmark performance to include a broader understanding of model capabilities and limitations. Therefore, we acknowledge that the in-depth exploration of these advanced prompting strategies represents a promising avenue for future studies.

\subsection{Scope of the study}
This paper's benchmarking parts are focused on assessing multiple-choice questions from quantitative reasoning and verbal reasoning sections of the GMAT, which form a significant but not complete portion of this exam. As explained in the methodology section, the GMAT includes 12 integrated reasoning (IR) and one essay section as well that are scored independently and their scores will not affect the exam's total score, so we did not include quantitative metrics for these two sections in our benchmarks. Consequently, the performance metrics we report may not fully reflect the LLMs capabilities in a real GMAT exam setting. Furthermore, while a study of LLMs on the GMAT can offer important data on certain cognitive abilities relevant to business education, it is not sufficient to generalize its findings to the entire field of business education without additional research that considers the full spectrum of skills and learning experiences that business programs aim to develop. Business education encompasses case studies, real-world problem-solving, interpersonal skills, and ethical decision-making, which may not be adequately captured by LLMs' performance on standardized tests. Additionally, success on the GMAT does not necessarily equate to success in business school or in business practice. The ability to transfer and apply test-taking skills to real-world scenarios is a critical aspect of business education that may not be reflected in the study's scope. Moreover, business schools often use a holistic approach to evaluate candidates, considering work experience, leadership potential, and other personal attributes alongside test scores. A study focused on GMAT performance alone may not account for these broader evaluative criteria.

\subsection{Memorization}
Table \ref{tab:Freemium} presents the performance of various models on both free and premium GMAT exams. It's important to note that accessing free exams requires users to create an account on the official GMAT exam website. Furthermore, to access premium exams, users must purchase them. When comparing the outcomes of free versus premium exams across all models, we observe no significant difference in performance. This finding, coupled with the fact that our GMAT materials come from official sources requiring login or payment for premium content, suggests that such content likely wasn't part of the LLMs' training data. Even if some overlap or contamination exists, it doesn't appear to notably enhance the LLMs' GMAT performance. OpenAI's research supports this, indicating that despite some contamination in publicly available benchmarks, it hasn't led to significant performance discrepancies between contaminated and uncontaminated samples in their assessments \cite{openai2023gpt4}.

\begin{table}[H]
    \centering
    \begin{tabular}{l >{\centering\arraybackslash}p{1.3cm} >{\centering\arraybackslash}p{1.3cm} >{\centering\arraybackslash}p{1.3cm} >{\centering\arraybackslash}p{1.5cm} >{\centering\arraybackslash}p{1.5cm} >{\centering\arraybackslash}p{1.75cm} >{\centering\arraybackslash}p{2cm}}
    \hline
    \toprule
    GMAT& \textbf {GPT-4 Turbo} &  GPT-4 &  GPT-3.5 Turbo& Claude 2.1&  Claude 2&  Gemini Pro (New Bard)& PaLM 2 (Legacy Bard)
\\\hline
Free Exam 1& \textbf {88.06}&  74.63&  61.19&  73.13&  62.69&  58.21& 43.28
\\
Free Exam 2& \textbf {79.10}&  71.64&  49.25&  73.13&  59.70&  77.61& 58.21
\\
Premium Exam& \textbf {88.06}&  76.12&  62.69&  70.15&  58.21&  76.12& 50.75
\\
\bottomrule
\textbf{Overall Average}& \textbf {85.07}&  74.13&  58.21&  72.14&  60.20&  70.65& 50.75\\
\bottomrule
    \end{tabular}
    \caption{Comparison of performance of models on the GMAT exams shows that GPT-4 Turbo significantly surpasses the other models.}
    \label{tab:Freemium}
\end{table}

\section{AI as Tutoring Assistant: A Case Study}
\begin{figure}
    \centering
    \includegraphics[width=1\linewidth]{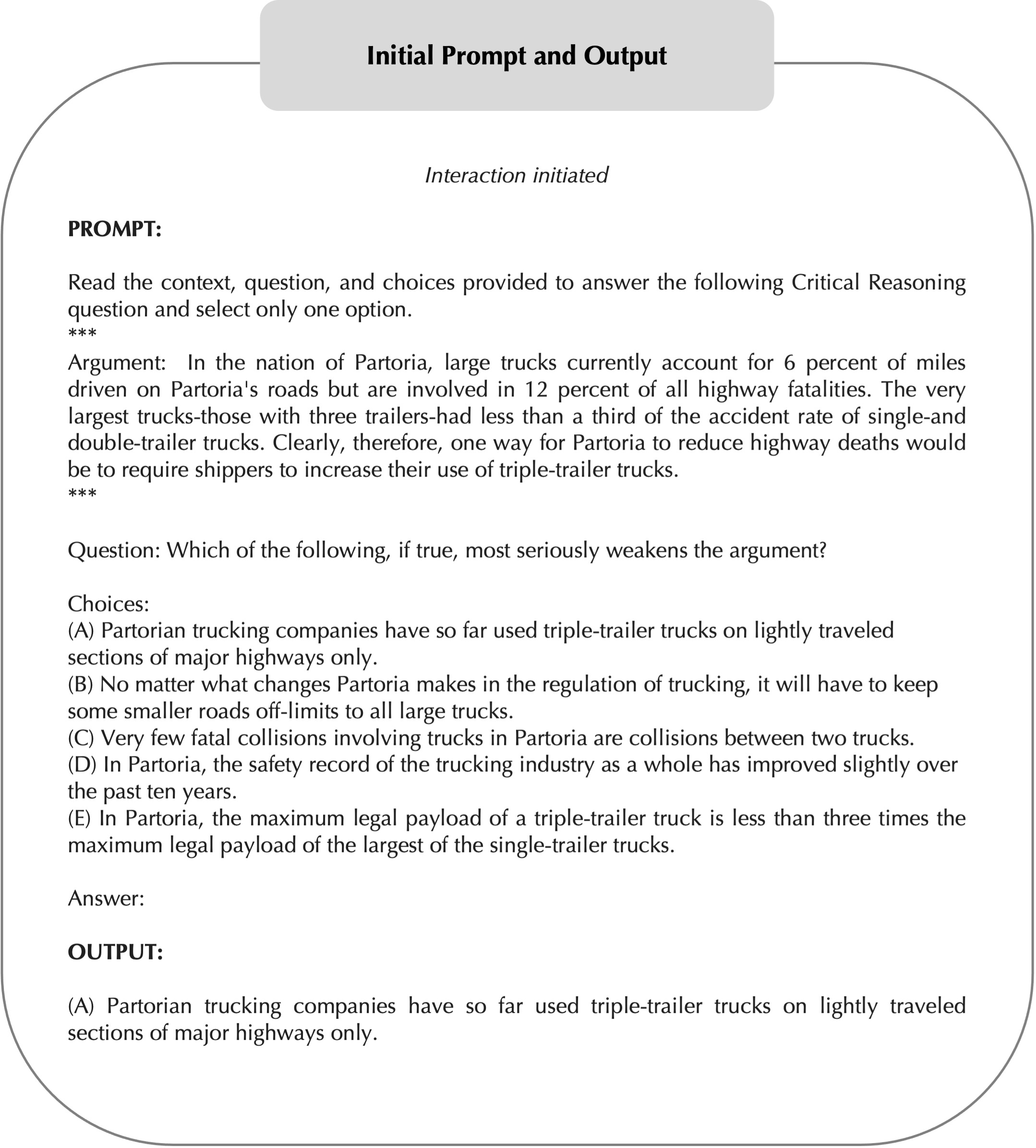}
    \caption{GPT-4 Turbo selects the correct option in response to a critical reasoning problem.}
    \label{fig:initial}
\end{figure}

\begin{figure}
    \centering
    \includegraphics[width=1\linewidth]{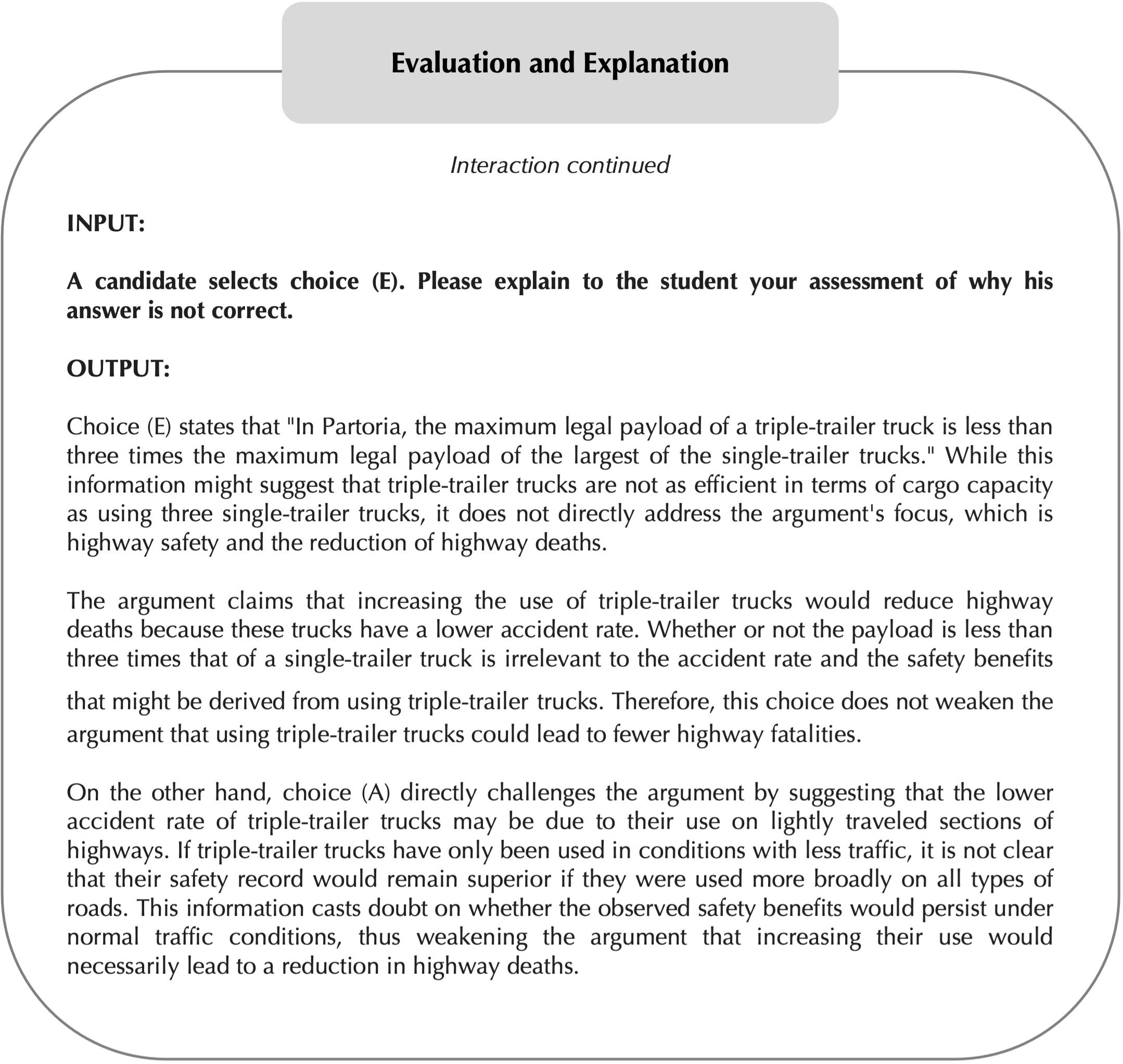}
    \caption{GPT-4 Turbo evaluates a candidate's response to the question and explains the rationale for the correct response.}
    \label{fig:ee}
\end{figure}

\begin{figure}
    \centering
    \includegraphics[width=1\linewidth]{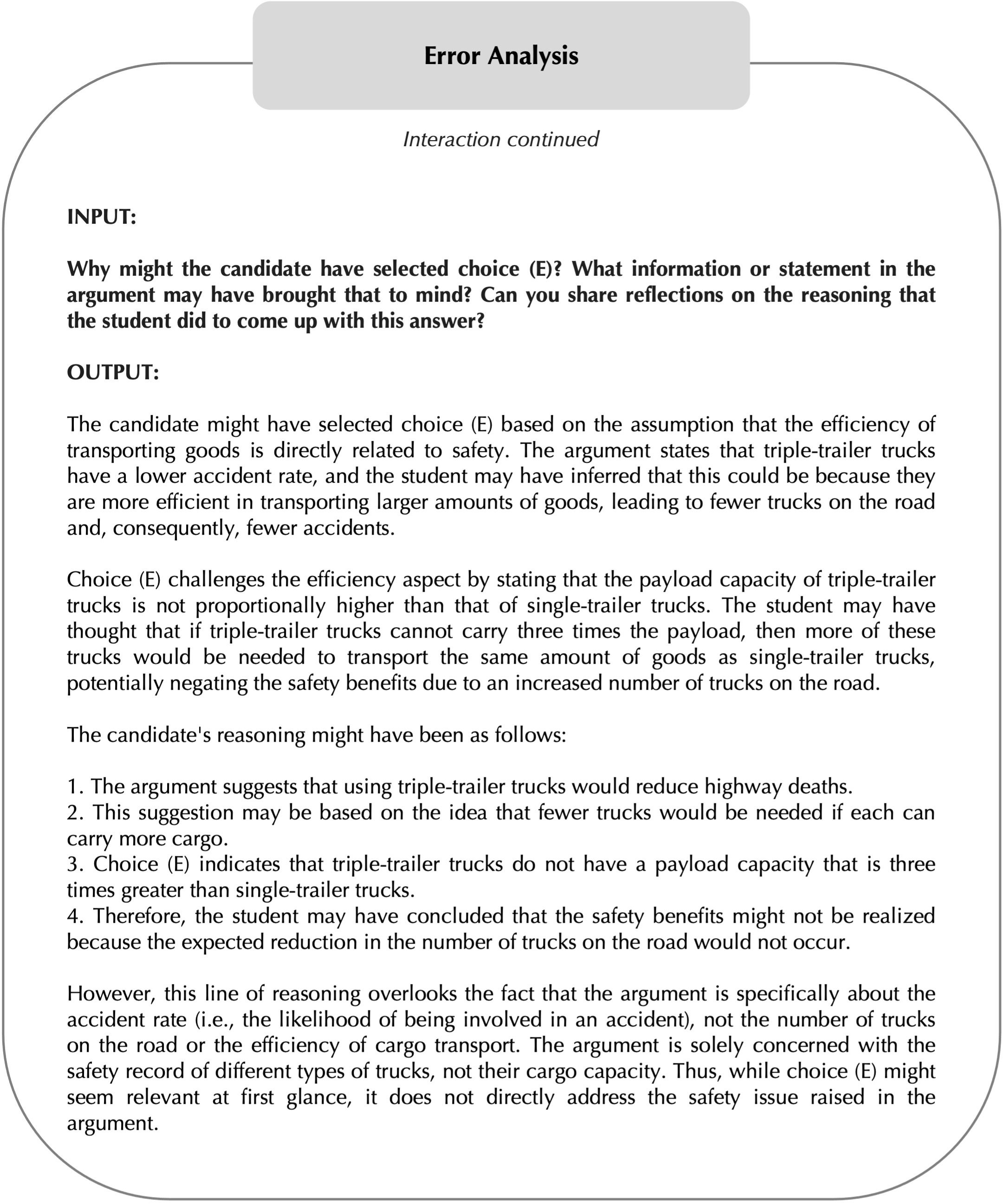}
    \caption{GPT-4 Turbo attempts to follow the candidate's line of reasoning to identify errors in his thought process.}
    \label{fig:ea}
\end{figure}

\begin{figure}
    \centering
    \includegraphics[width=1\linewidth]{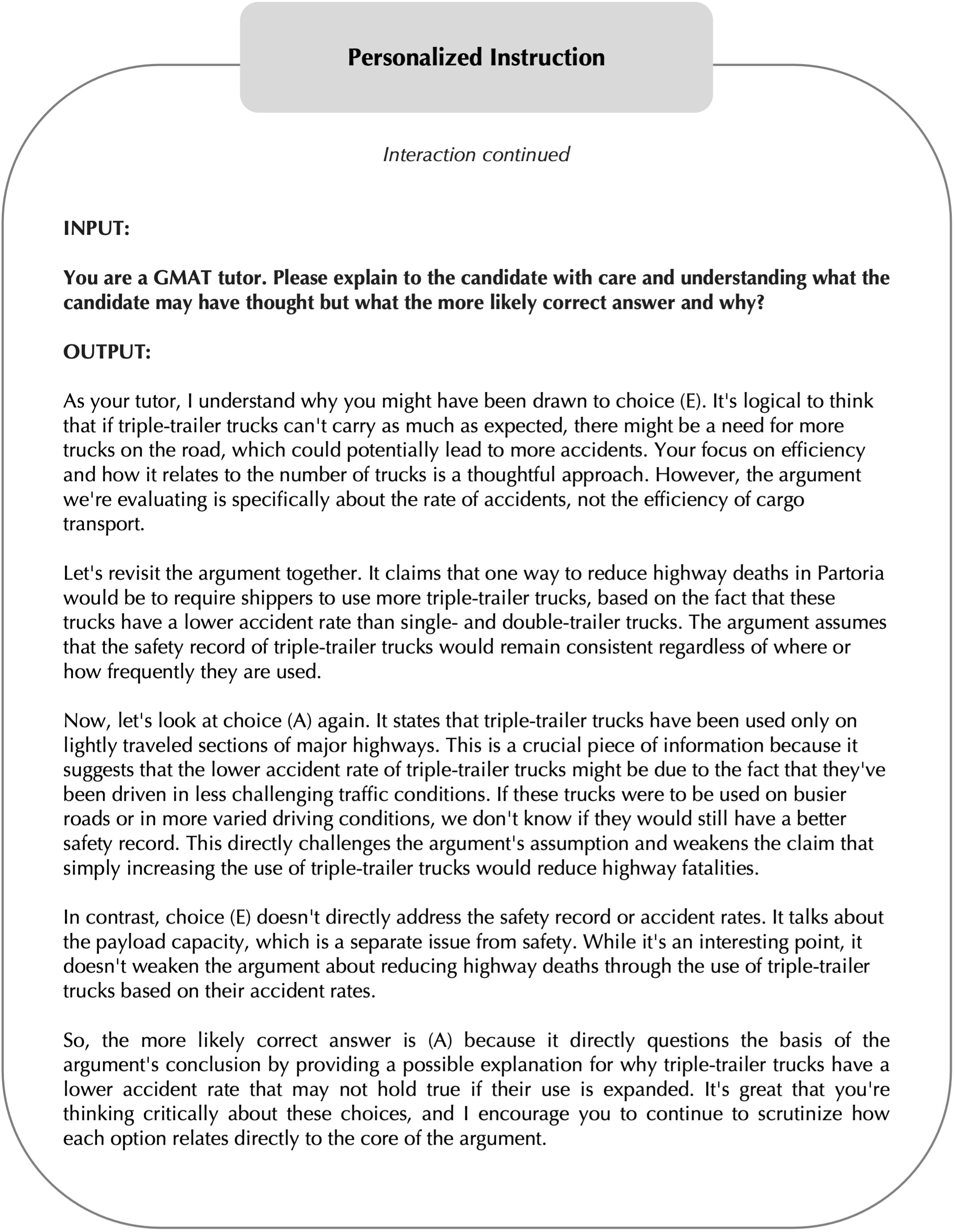}
    \caption{GPT-4 Turbo adjusts its tutoring approach for personalized instruction.}
    \label{fig:pi}
\end{figure}

\begin{figure}
    \centering
    \includegraphics[width=1\linewidth]{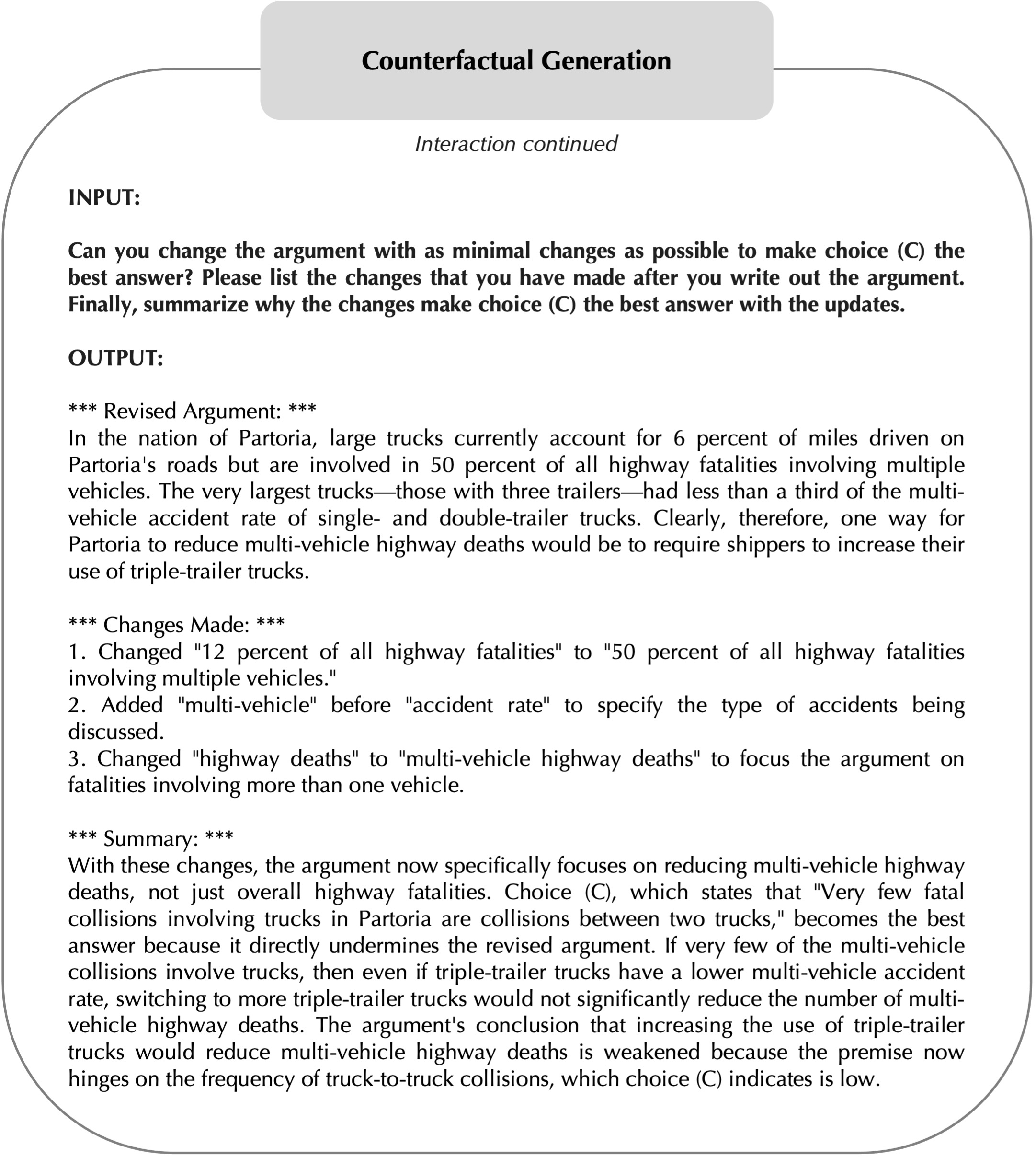}
    \caption{GPT-4 Turbo revises the argument in the problem statement in order to construct a counterfactual scenario.}
    \label{fig:cg}
\end{figure}

LLMs can act as tutoring assistants, simplifying complex concepts for both instructors and students. They can guide students through homework by fostering critical thinking instead of just providing answers. LLMs can generate practice questions, offer assignment feedback, and create tailored study plans. They're also adept at simulating exams, tracking progress, and suggesting targeted practice areas, especially useful for language exams that focus on vocabulary and grammar. Additionally, LLMs help with essay writing and provide motivational support. As AI becomes more integrated into education, it offers interactive and personalized learning experiences. Engaging LLMs in interactive sessions can reveal their educational potential and practical applications. Our study, which references the interactive session described in \cite{nori2023capabilities}, illustrates how a dialogue initiated by a single critical reasoning question can showcase an AI model's educational capabilities. Using GPT-4 Turbo, we simulate a conversation between the model and a student preparing for the GMAT, demonstrating the model's ability to correctly answer questions (Figure \ref{fig:initial}), explain the reasoning behind answers (Figure \ref{fig:ee}), use theory-of-mind to hypothesize about errors (Figure \ref{fig:ea}), personalize instruction (Figure \ref{fig:pi}), and engage in counterfactual reasoning by modifying a critical reasoning problem to help the candidate consider different outcomes (Figure \ref{fig:cg}). However, it is essential to verify the accuracy of information generated in such interactions and in real-world applications with expert review to ensure reliability. This investigation aims to highlight AI's practical applications in education and its potential to facilitate personalized learning journeys.

\section{Final Thoughts}
The overall performance of Large Language Models (LLMs) on GMAT exam demonstrates significant potential to revolutionize education and tutoring in particular, while also highlighting certain challenges and limitations. We anticipate that LLMs will become essential tools for students in their exam preparation and for teachers in developing their pedagogical materials. Additionally, we expect these models to significantly enhance the process of standardized examinations. These insights suggest a rapidly evolving landscape where AI models are not only becoming viable candidates for high-level academic programs but are also setting new benchmarks for human applicants to aspire to. The implications for the future of standardized testing and admissions are profound, as AI continues to redefine the limits of performance and potential. Nevertheless, there are some major concerns regarding the deployment of LLMs in educational settings, specifically within the domain of business education, that warrant careful consideration.

\paragraph{Social impacts.}
One of the major concerns is that the use of LLMs in education could lead to increased inequality \cite{hannan2023ai}. Access to LLMs requires a stable internet connection and devices capable of running such models, which may not be available to all students. Therefore, students with access to these technologies might gain unfair advantages over those who do not, potentially exacerbating existing educational inequalities. Also, several studies indicate that LLMs may perpetuate and amplify existing biases present in their training data, potentially reinforcing stereotypes and contributing to inequality in educational outcomes. Moreover, while research shows that face-to-face interaction plays a crucial role in providing a sustainable learning experience in business education \cite{fish2023comparing}, overreliance on LLMs for learning could lead to a decrease in face-to-face interactions and the valuable social aspects of learning, such as collaboration and discussion. Furthermore, the adoption of LLMs could lead to concerns about job displacement \cite{kiradoo2023unlocking}, as AI might be seen as a replacement for human teachers and tutors.

\paragraph{Risk of misinformation.}
As our findings in this study showed, LLMs may generate incorrect or misleading information, which could lead to confusion or the learning of inaccurate material.  As a result, students may inadvertently learn and propagate these inaccuracies, leading to a spread of misinformation. 

\paragraph{Impacts on personal development.}
LLMs might provide answers too readily, which could discourage students from developing their own critical thinking, problem-solving skills, and also interpersonal skills. As a result, students might become dependent on AI, leading to a decrease in self-efficacy and confidence in their abilities to learn and solve problems independently.

\paragraph{Ethical considerations.}
The integration of LLMs into education involves the collection and processing of student data, raising concerns about privacy, data security, and the potential misuse of personal information. Also, there may be concerns about the use of copyrighted material when LLMs generate content or examples for study purposes. Additionally, there is a risk that students might use LLMs to complete assignments dishonestly, leading to challenges in ensuring academic integrity and authentic assessment of student learning. 

To mitigate these concerns, it's important to use LLMs as a complement to traditional educational methods and human interaction in business education, rather than as a standalone solution. Additionally, ongoing monitoring, evaluation, and adjustment of LLMs in educational settings are necessary to ensure they are used effectively and ethically and to ensure that students are developing their own critical thinking and problem-solving skills, rather than becoming overly reliant on the technology. Furthermore, to regulate and monitor the ethical use of AI in education, it is essential to establish frameworks, guidelines, and oversight bodies.

Finally, while LLMs have demonstrated their implications in a wide range of applications, they have several inherent limitations. Current LLMs, trained on vast internet text sources like Reddit and Twitter, have versatile capabilities like scriptwriting, translation, and recipe generation, but they often stray off-topic, fabricate information, and show fluidity in perspectives. Furthermore, while highly capable in strictly defined calculations, current limitations persist in mathematical ambiguity and complexity intrinsic to human cognition. Extending their precise computational strengths while approximating human understanding and fluidity thus remains an elusive frontier. LLMs face challenges with abstract or intricate concepts that demand reasoning beyond the literal text. Their performance dips when tasks require creative problem-solving, intuitive judgment, or comprehension of context and linguistic nuances. Additionally, they may inadvertently change the original meaning of a sentence due to their limited understanding of the writer's intent or the deeper connotations of the information. OpenAI tries to mitigate some of these issues by fine-tuning LLMs with human feedback and reinforcement learning \cite{lambert2022illustrating}, aiming for helpful, truthful, and harmless outputs, though these goals can be subjective and challenging to define. The statistical models of crowdsourcing, introduced by Phil Dawid and David Skene in 1979 \cite{dawid1979maximum} and expanded upon in Cultural Consensus Theory \cite{romney1986culture} (CCT) by Romney and colleagues, offer insights into assessing subjective judgments. CCT, particularly, recognizes the lack of a singular consensus in diverse opinions, a concept valuable for understanding and improving LLMs in handling varied, complex information. This principle has led to further advancements, including the fine-tuning of deep neural networks to align their outputs more closely with consensus beliefs.\cite{gurkan2023harnessing}

\section{Conclusion}
In our study, we performed a comparative analysis of seven general-purpose LLMs—GPT-4 Turbo, GPT-4, GPT-3.5 Turbo, Claude 2.1, Claude 2, Gemini 1.0 Pro, and PaLM 2—focusing on their zero-shot performance in the GMAT's quantitative and verbal sections, which are crucial for global business school admissions. GPT-4 Turbo stood out as the top performer, with the latest versions of the other models also showing significant improvements over their predecessors. We found that LLMs can achieve impressive results on the GMAT without the need for complex prompting strategies, surpassing the average human test-taker. This suggests that LLMs have inherent strengths that make them well-suited for standardized testing. Our research provides a foundational understanding of LLMs' capabilities on the GMAT and suggests potential for further performance optimization. Additionally, we showcased GPT-4 Turbo's tutoring abilities, such as explaining critical reasoning, assessing responses, using metacognition to pinpoint student errors, and creating counterfactual scenarios. We explored the broader implications of AI in educational settings, suggesting that LLMs' strong performance on the GMAT indicates their potential value in business education and as a support tool for business professionals. However, the potential for errors and the complexities of real-world application call for caution. It is essential to carefully develop, evaluate, and refine AI use, and to continue technological advancements to maximize the benefits and minimize the risks of AI in education.

\bibliographystyle{unsrt}  

\newpage

\appendix
\renewcommand\thefigure{\thesection.\arabic{figure}}
\setcounter{figure}{0} 
\section{Sample of Input/Output for various GMAT tasks}
\begin{figure}[ht]
    \centering
    \includegraphics[width=\linewidth]{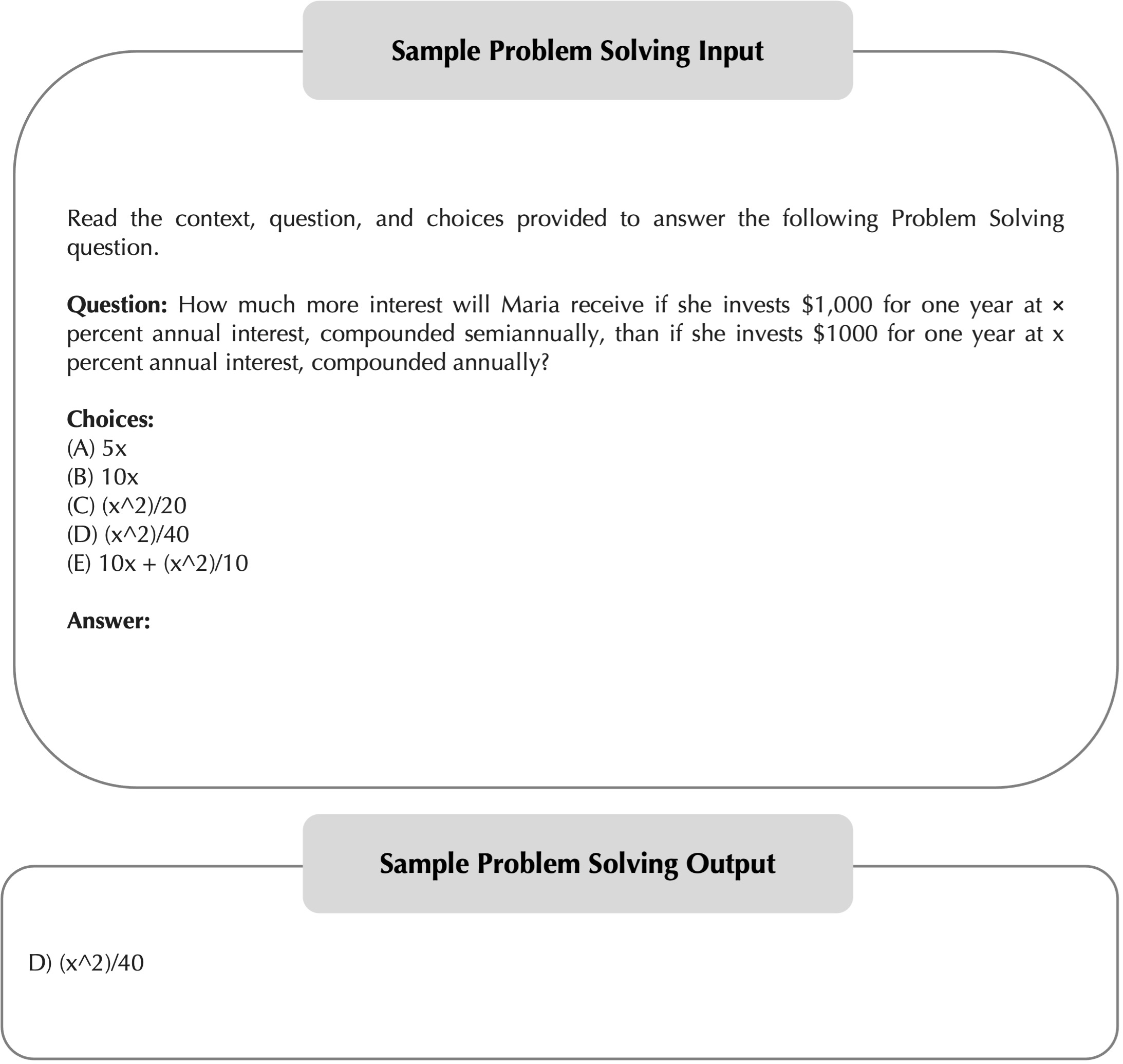}
    \caption{Sample of the user's input and the model's response for a problem solving task}
    \label{fig:ps}
\end{figure}

\newpage
\begin{figure}[ht]
    \centering
    \includegraphics[width=1\linewidth]{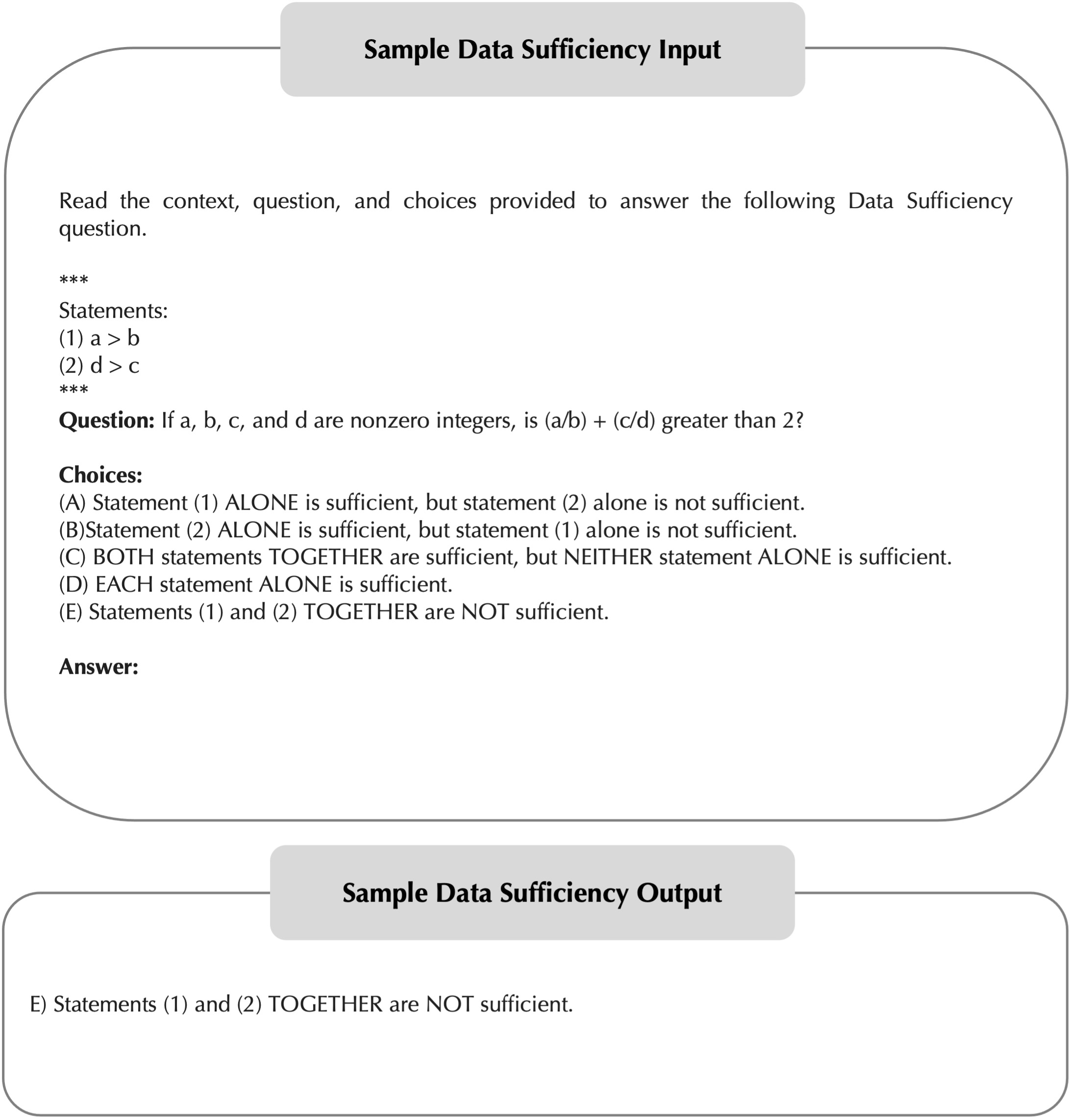}
    \caption{Sample of the user's input and the model's response for a data sufficiency task }
    \label{fig:enter-label}
\end{figure}

\newpage
\begin{figure}[ht]
    \centering
    \includegraphics[width=1\linewidth]{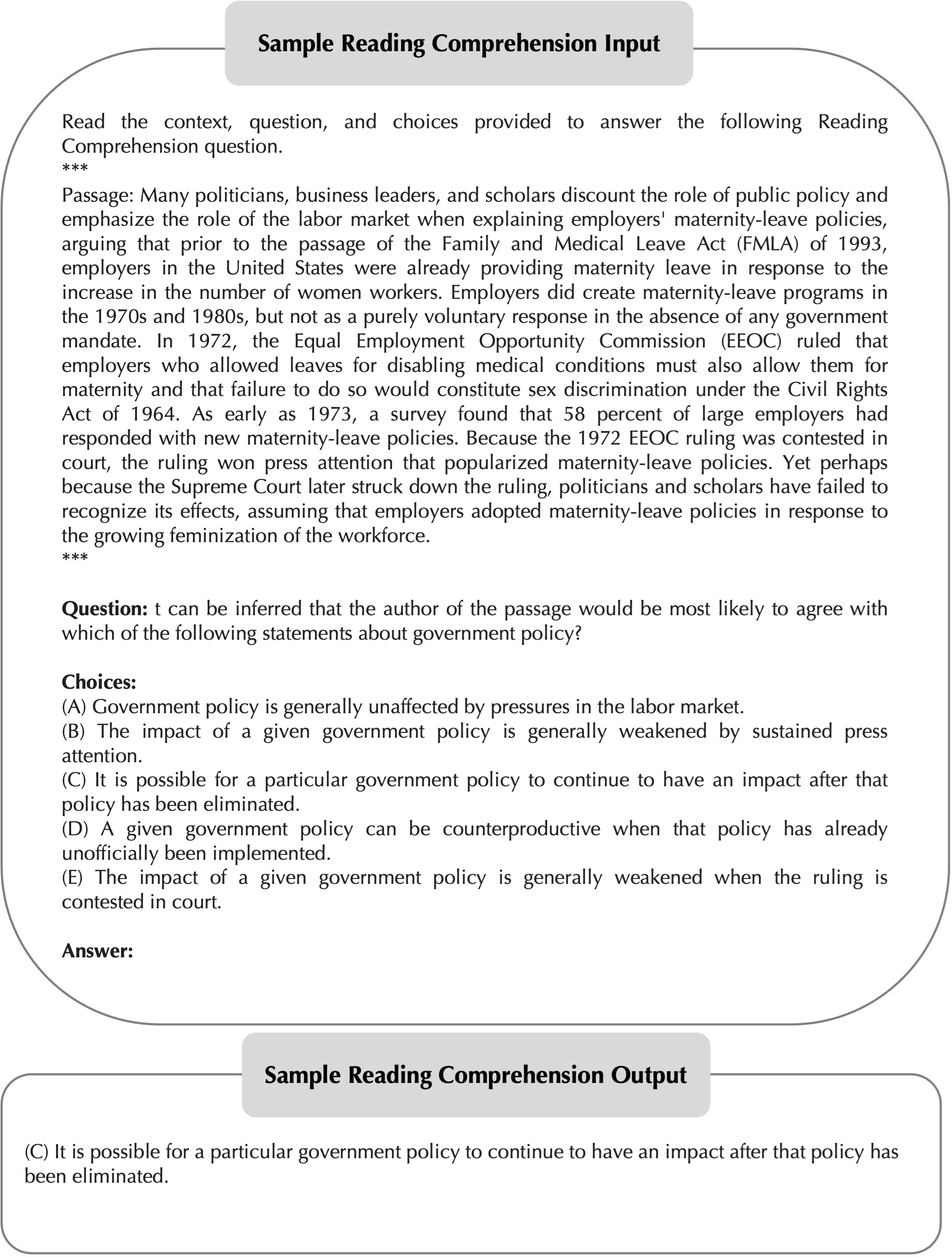}
    \caption{Sample of the user's input and the model's response for a reading comprehension task }
    \label{fig:rc}
\end{figure}
\newpage
\begin{figure}[ht]
    \centering
    \includegraphics[width=1\linewidth]{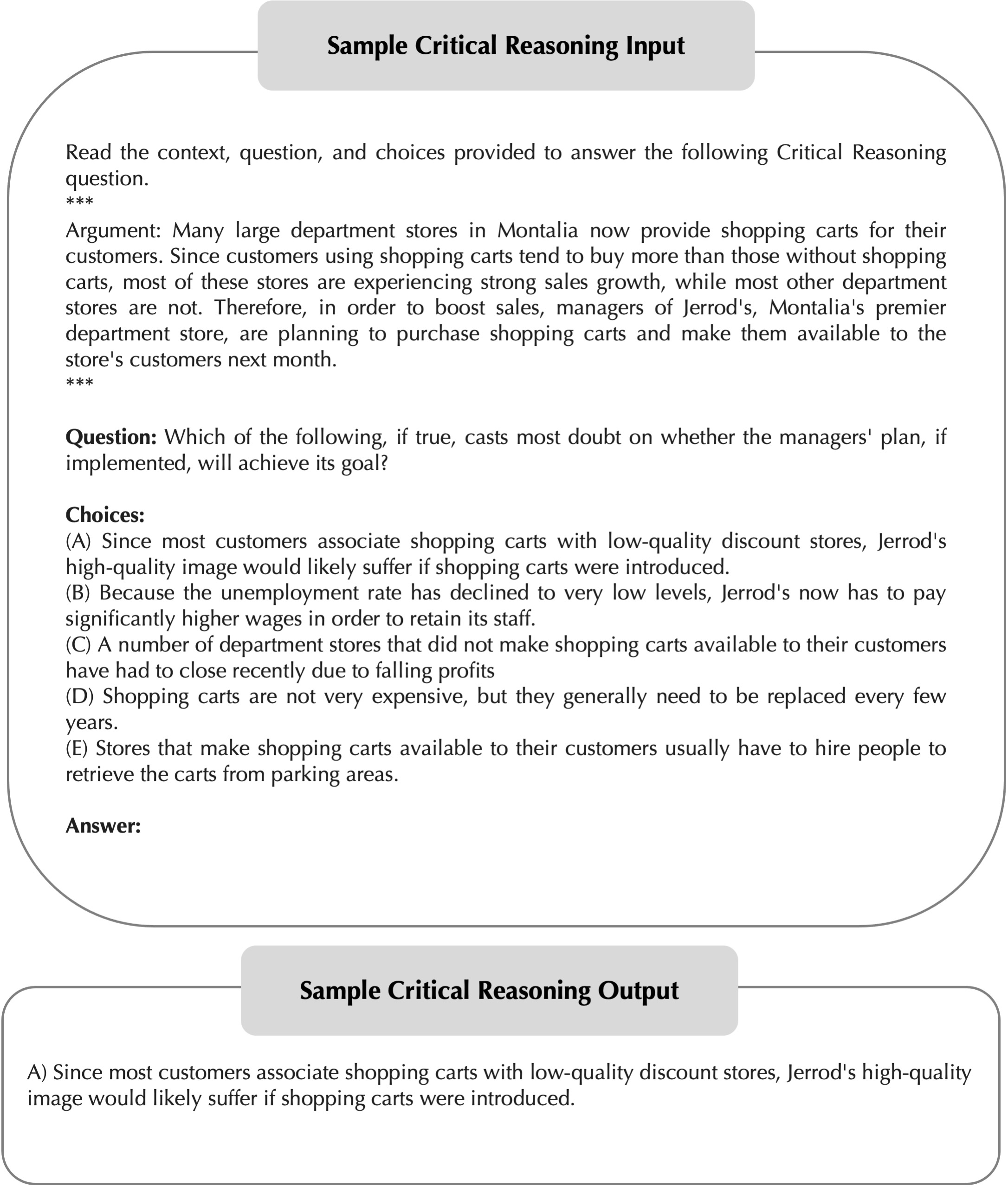}
    \caption{Sample of the user's input and the model's response for a critical reasoning task }
    \label{fig:cr}
\end{figure}
\begin{figure}[ht]
    \centering
    \includegraphics[width=1\linewidth]{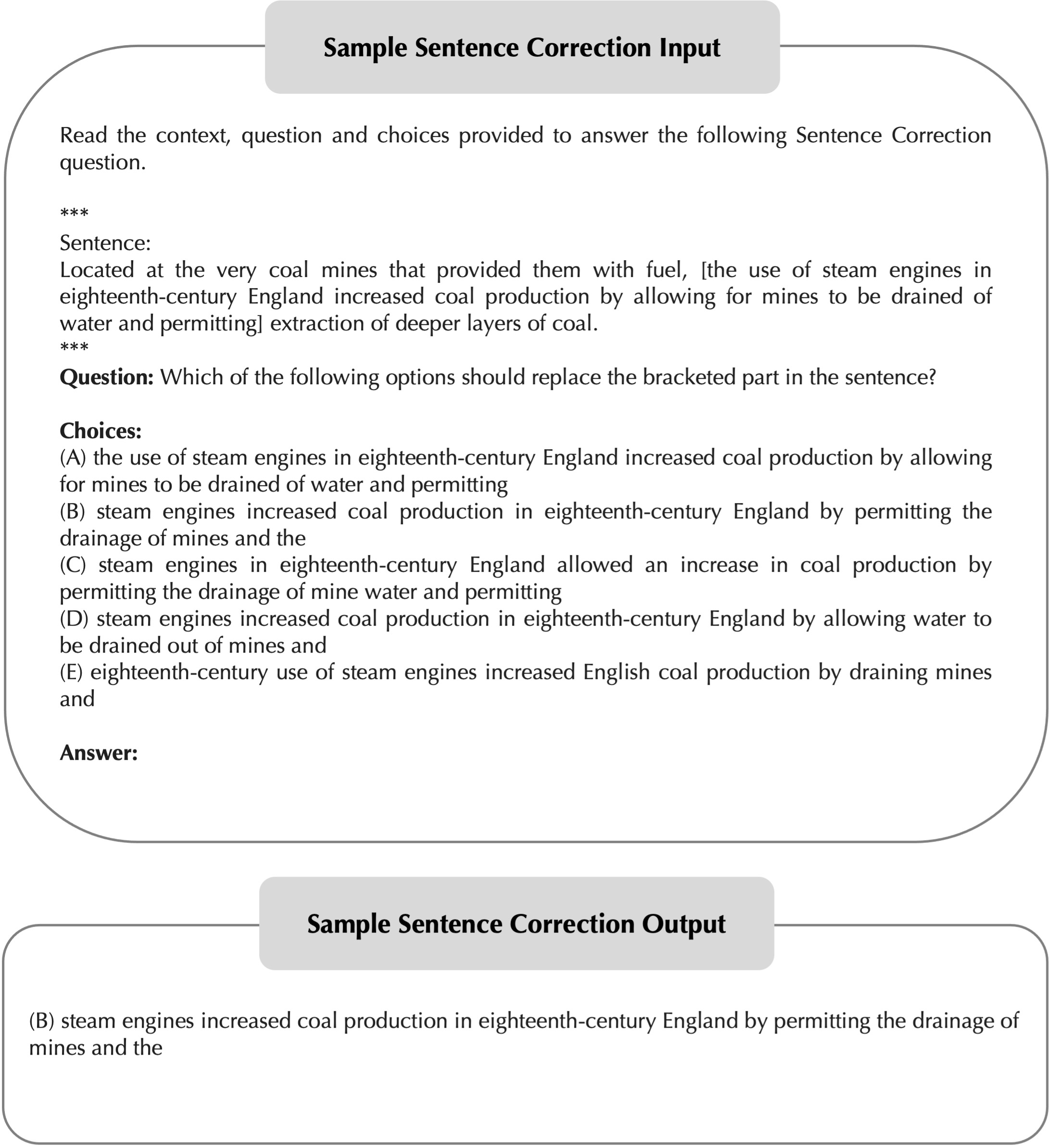}
    \caption{Sample of the user's input and the model's response for a sentence correction task}
    \label{fig:sc}
\end{figure}

\end{document}